\documentclass[11pt]{article}
\usepackage[utf8]{inputenc}
\usepackage{CJKutf8}

\usepackage[final]{acl}

\usepackage{times}
\usepackage{latexsym}
\usepackage{fix-cm}

\usepackage[T1]{fontenc}

\usepackage[utf8]{inputenc}

\usepackage{newunicodechar}
\newunicodechar{（}{(}
\newunicodechar{）}{)}
\usepackage{microtype}

\usepackage{inconsolata}

\usepackage{graphicx}
\usepackage{comment}

\usepackage{booktabs}
\usepackage{multirow}
\usepackage{amsmath}
\usepackage{amssymb}  %
\usepackage{enumitem}
\usepackage{float}
\usepackage{tikz}
\usepackage{tablefootnote}
\usepackage{textcomp}
\usepackage[safe]{tipa}
\newcommand{\dotB}{\tikz[baseline=-0.55ex]\fill[blue!75!black] circle (2.6pt);}
\newcommand{\dotR}{\tikz[baseline=-0.55ex]\fill[red!75!black]  circle (2.6pt);}
\newcommand{\dotG}{\tikz[baseline=-0.55ex]\fill[green!55!black] circle (2.6pt);}
\usepackage{amssymb}
\usetikzlibrary{arrows.meta}
\usepackage{hyperref}
\usepackage{array}
\usepackage{siunitx}

\newcommand{\emb}[1]{\mathbf{e}_{#1}}

\usepackage{subcaption} %

\title{Measuring Data Diversity with Embeddings: \\ A Taxonomy and Benchmark for Diversity Measures}

 \author{Cantao Su, Anna Wegmann, Esther Ploeger, Dong Nguyen \\
         Department of Information \& Computing Sciences, Utrecht University \\Utrecht, The Netherlands \\
         \small{
   \textbf{Correspondence:} \href{mailto:c.su@uu.nl}{c.su@uu.nl}
 }}

\begin{document}
\maketitle
\begin{abstract}
Diversity is a critical property of text corpora, with direct implications for training data curation, test set design, and model generalization. However, its measurement remains fragmented and poorly understood. 
Existing embedding-based diversity measures vary widely in how they are computed and can yield substantially different diversity scores on the same data, yet no systematic comparison of their empirical behaviour exists. 
We make two contributions toward addressing this gap. First, we present a taxonomy of 21 embedding-based diversity measures, organizing them into four families based on their underlying computations.
Second, we introduce \textbf{EmbDivBench}, a benchmark that evaluates these measures along three conceptually distinct components---variety, balance and disparity---on both simulated and natural text data. 
Although Mean Pairwise Distance is the most commonly used measure, our experiments show that it is outperformed by several alternatives on the three diversity components, with Graph Entropy being the best choice overall. Lastly, we distil these findings into practical recommendations for diversity measure selection.
\end{abstract}

\section{Introduction} \label{sec:introduction}

Data diversity has demonstrated practical value across multiple areas of NLP. 
First, diversity in training data is a driver of model capability: corpora that are more diverse have shown performance improvement across a range of tasks \citep{liuWhyCantDiscourse2023, yadav_explicit_2024, chenDiversitySyntheticData2024}, %
and can compensate for smaller data sizes \citep{schillerDiversitySizeEffect2024, esteveDiversityDietHealthier2026}.
Second, %
it helps to improve fairness and inclusiveness \citep{blodgettTwitterUniversalDependency2018}, %
counteract the dominance of English- and Western-centric data \citep{pouran_ben_veyseh_minion_2022}, %
and reduce toxic output and bias \citep{chuFairnessLargeLanguage2024, agarwalSafeTextUnifiedApproach2026, gehman_realtoxicityprompts_2020}.
Third, diversity is such an intrinsic property of human language that generating  sufficiently diverse outputs is itself a user expectation \citep{esteveSurveyDiversityQuantification2025, leeRandomizedLinkTransformer2022}.

\begin{figure}[!t]
  \centering
  \includegraphics[width=\columnwidth]{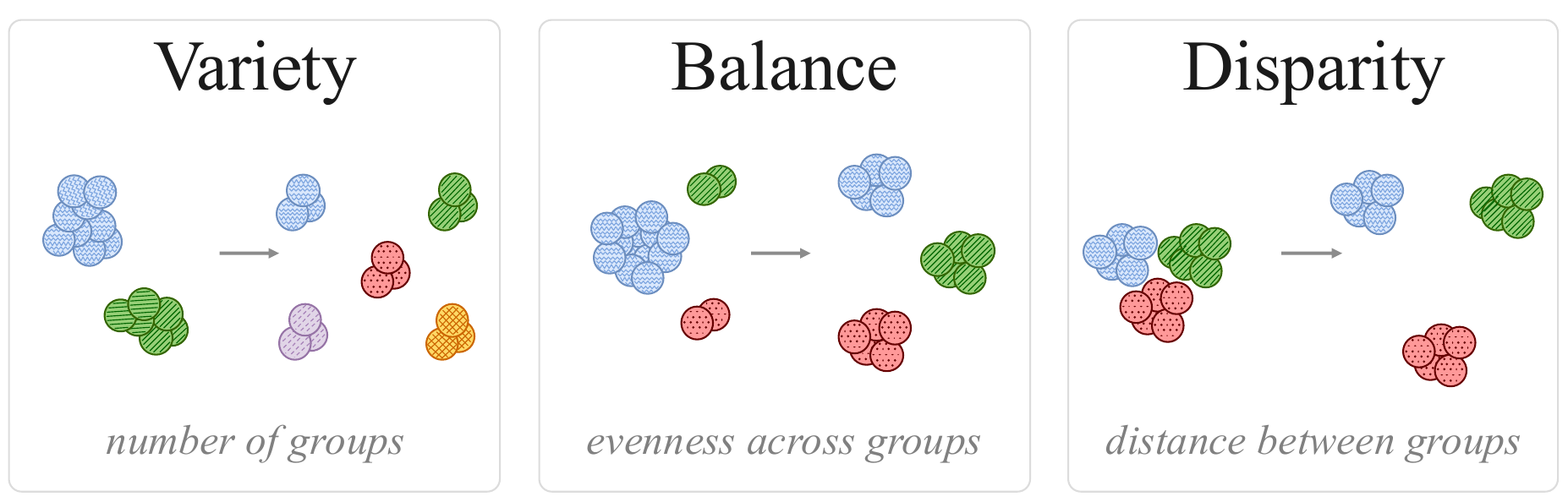}
  \caption{The three components of diversity in \textsc{EmbDivBench}.
    In each panel, the left configuration has lower diversity and
    the right has higher diversity along the corresponding component.
    \textbf{Variety} counts the number of groups present;
    \textbf{Balance} measures how evenly points are distributed
    across groups; \textbf{Disparity} captures how far apart
    the groups lie in the embedding space.}
  \label{fig:vbd-axes}
\end{figure}

Despite its importance, the measurement of diversity has not kept pace. First, ``diversity'' remains a vague and overloaded term in NLP \citep{nguyen-ploeger-2025-need, esteveSurveyDiversityQuantification2025}:
Datasets are described as ``diverse'' without specifying what 
\textit{dimension} (e.g., stylistic vs. semantic) and \textit{component} (e.g., variety vs. balance in Fig.~\ref{fig:vbd-axes}) of diversity is meant \citep{zhaoPositionMeasureDataset2024, shaibStandardizingMeasurementText2025, guoCuriousDeclineLinguistic2024,nguyen-ploeger-2025-need}.
Outside NLP, diversity has long been decomposed into three distinct components (Fig.~\ref{fig:vbd-axes}, see also \S\ref{subsec:axes})---variety (how many categories the data spans), balance (how evenly data are distributed across them), and disparity (how different the categories are from one another) \citep{stirlingGeneralFrameworkAnalysing2007}. 
These components have more recently been used in NLP, for example, to theoretically categorize measures into what components they are likely to be sensitive to \citep{esteveSurveyDiversityQuantification2025}.
However, different measures have been named and computed inconsistently (\S\ref{sec:taxonomy:landscape}) and used in isolation %
without an overview organizing the differences in how they are computed, or how they relate. We introduce a unifying taxonomy of existing embedding-based diversity measures based on how they are computed (\S\ref{sec:taxonomy}).

Second, despite the increasing use of diversity measures, the measurements themselves are rarely evaluated, with a few exceptions (e.g., \citealt{tevetEvaluatingEvaluationDiversity2021,yangMeasuringDataDiversity2025,zhang2025evaluating}). However, systematic evaluation is crucial: otherwise, it remains unclear which measure researchers should use and whether newly proposed measures truly improve over existing ones.
We introduce \textsc{Emb\allowbreak DivBench}, an empirical benchmark that, contrary to previous approaches (\S\ref{sec:related}), tests sensitivity to variety, balance, and disparity through controlled manipulations of each component. %

Among approaches to diversity measurement, embedding-based methods have become increasingly popular \citep{shaibStandardizingMeasurementText2025, guoBenchmarkingLinguisticDiversity, zhuMeasuringDiversitySynthetic2025, yangMeasuringDataDiversity2025, fanCOMBATTINGDIMENSIONALCOLLAPSE2025, friedman2022vendi}.
The success of representation learning in NLP %
 makes it possible to represent different dimensions (e.g., semantic or stylistic) of text as points in a continuous vector space \cite{reimersSentenceBERTSentenceEmbeddings2019,wegmann-etal-2022-author}. %
Embedding-based diversity measures are thus widely applicable: diversity can be measured along many dimensions, as long as texts can be mapped to an embedding space capturing the relevant diversity dimension.
Although some studies have compared diversity measures \cite{esteveSurveyDiversityQuantification2025,yangMeasuringDataDiversity2025}, none have provided a holistic categorization or an empirical analysis of embedding-based diversity measures. To address these gaps,    
we make the following \textbf{contributions}:
\begin{itemize}[itemsep=0.3em, parsep=0pt, topsep=1pt]
    \item A \textbf{taxonomy} (\S\ref{taxonomy}) that surveys and classifies \textbf{21}
    embedding-based diversity measures into four organized families.%
    \item \textbf{\textsc{EmbDivBench}} (\S\ref{sec:benchmark}), a benchmark that %
    tests measures on the variety, balance and disparity components through two controlled complementary tiers: a simulated tier that affords exact control over each component, and a natural text tier for the semantic dimension. %
    \item \textbf{Practical recommendations} (\S\ref{recommendations}) for diversity measure selection, on each component. %
    \item \textbf{Data and code} are publicly available on GitHub\footnote{\url{https://github.com/nlpsoc/EmbDivBench}} and can be used to compare new diversity measures. 
\end{itemize}

    We find that disparity is the hardest component for diversity measures and that natural text is more difficult for measures than the simulated data tier. 
    Mean Pairwise Distance---a popular embedding-based diversity measure---is outperformed by several alternatives on all three components, with Graph Entropy performing the best overall.

\section{Related Work}
\label{sec:related}
 
\paragraph{Non-embedding-based diversity measures.}
Diversity has been studied across many fields---e.g., in ecology, economics, and sociology. In many of these fields, Shannon Entropy is a popular approach \citep{leinsterMeasuringDiversityImportance2012, nehringTheoryDiversity2002}. 
Within NLP, the dominant non-embedding measures are usually surface-form statistics over tokens or $n$-grams, such as type-token ratio and its variants \cite{covingtonCuttingGordianKnot2010}, Distinct-$n$ \citep{liDiversityPromotingObjectiveFunction2016a}, Self-BLEU \citep{zhuTexygenBenchmarkingPlatform2018}, and $n$-gram entropy \cite{hanMeasuringImprovingSemantic2022}. 
While useful, they only capture surface variation \citep{stasaski-hearst-2022-semantic, tevetEvaluatingEvaluationDiversity2021}. Such non-embedding-based measures have also been included in toolkits, to apply them over dependency-parsed corpora \citep{esteve-dobrovoljc-2026-delta} or between language pairs \citep{goot-etal-2025-distals}.
In contrast, we focus on embedding-based diversity measurement---a promising and increasingly popular way to measure data diversity. %

\paragraph{Embedding-based diversity measures: many but scattered.}
Advances in representation learning have made it possible to encode discrete objects---sentences, images, molecules---as points in a continuous space whose structure reflects similarity, opening up a class of diversity measures that operate directly on these representations rather than on surface form or predefined categories. Within NLP, a growing number of works exploit this: Mean Pairwise Distance has been used most frequently \citep{tevetEvaluatingEvaluationDiversity2021,shaibStandardizingMeasurementText2025,mirandaScaleDiversityCoefficient2025,ploegerWhatTypologicalDiversity2024}. 
Other measures used in NLP include Cluster Inertia \citep{duBoostingDialogResponse2019,yangMeasuringDataDiversity2025,zhuMeasuringDiversitySynthetic2025}, the geometric mean of per-dimension standard deviations across all embeddings \citep{laiDiversityDensityHomogeneity2020, wangDiversityorientedDataAugmentation2025, yangMeasuringDataDiversity2025}, the Vendi Score \citep{friedman2022vendi, yangMeasuringDataDiversity2025, zhang-etal-2024-improving-diversity, baPredictTrainingData2025}, and DCScore \citep{zhuMeasuringDiversitySynthetic2025, baMeasuringDatasetDiversity2026}. Across this literature, these measures are largely introduced and used as general-purpose diversity measures, without systematically distinguishing the different components of diversity they may capture.

Embedding-based measures have also been proposed and used in other fields, such as 
Hamiltonian Diversity in chemistry \citep{huHamiltonianDiversityEffectively2024, duMolGenSurveySystematicSurvey}, %
Energy over graph-distance embeddings for graph generation \citep{velikonivtsevChallengesGeneratingStructurally2024}, %
and Rényi Kernel Entropy in computer vision \citep{jalali2023information}. %
Because some measures could also be applied to text, we also include them in our taxonomy and test them on our benchmark. An overview of all included measures is in App.~\S\ref{app:taxonomy}.%

\paragraph{Evaluation of diversity measures.}
In most of the work above, diversity measures serve as a tool rather than as the object of study.
Diversity measurement itself has received limited systematic evaluation, and researchers have recently called for more attention to this issue \citep{yangMeasuringDataDiversity2025,zhaoPositionMeasureDataset2024,nguyen-ploeger-2025-need}.
Prior work has evaluated measures by assessing their scores on text sets of known 
diversity level, e.g., by varying the temperature parameter \citep{tevetEvaluatingEvaluationDiversity2021,zhuMeasuringDiversitySynthetic2025}, the number of paraphrases \citep{zhang2025evaluating} and by asking crowd workers to produce text sets with higher or lower diversity \citep{tevetEvaluatingEvaluationDiversity2021}. Crucially, none of these evaluations distinguishes between the variety, balance and disparity components of diversity.

Diversity measures have also been correlated against human judgements of diversity \citep{hanMeasuringImprovingSemantic2022,zhuMeasuringDiversitySynthetic2025}.
However, eliciting human judgements of text diversity is challenging and costly, can be influenced by other factors like text quality, and is only feasible for small sets of texts \citep{zhang2025evaluating,tevetEvaluatingEvaluationDiversity2021}. A recent study by \citet{zhang2025evaluating} compares diversity measures against LLM-annotated judgements of diversity. While more scalable, such evaluations are dependent on the quality of the LLM annotations and factors like prompt design.

Diversity measures have also been evaluated by correlating them with performance on a downstream task  \citep{yangMeasuringDataDiversity2025}; however, such correlations do not necessarily demonstrate that the measure captures diversity, as performance gains may result from other factors captured by the measure.
Studies have also examined diversity measures by correlating them against each other \citep{zhang2025evaluating,shaibStandardizingMeasurementText2025}.

Most directly related to our work is \citet{esteveSurveyDiversityQuantification2025}, who survey diversity measurement in NLP and apply Stirling's variety, balance and disparity (V-B-D) decomposition to assign existing measures to the component their formula should target, without empirical testing.
Instead, our taxonomy classifies embedding-based measures by how the score is computed (\S\ref{sec:taxonomy}), and \mbox{\textsc{EmbDivBench}} tests their component sensitivity empirically (\S\ref{sec:benchmark}). To the best of our knowledge, this is the first empirical component-based evaluation of diversity measures in NLP.

\section{Unified Taxonomy of Diversity Measures}\label{taxonomy}

We describe the current fragmented landscape of embedding-based diversity measurement (\S\ref{sec:taxonomy:landscape}) and unify 21 measures into a taxonomy (\S\ref{sec:taxonomy}).

\subsection{A Fragmented Landscape}
\label{sec:taxonomy:landscape}

The literature on embedding-based diversity measurement is strikingly inconsistent in how it defines and uses measures. 
We collect 21 different embedding-based diversity measures (Tab.~\ref{tab:taxonomy}) and highlight three patterns of fragmentation that motivate the need for a unified taxonomy.

\paragraph{Different names, same computation.}
The same computation is sometimes introduced under multiple names by different researchers. For example, the sum of pairwise distances over a dataset is called \textit{Max Dispersion} by \citet{yuCanDataDiversity2022},  \textit{DistSum} by \citet{yangMeasuringDataDiversity2025} and \textit{Remote-Clique} by \citet{cevallosDiversityMaximizationDoubling2018}. %
Another example is that the geometric mean of per-dimension standard deviations is called \textit{Diversity} by \citet{laiDiversityDensityHomogeneity2020} but \textit{Radius} by \citet{yangMeasuringDataDiversity2025}.

\paragraph{Same name, different computations.}
Conversely, the same name is sometimes used for distinct computations. The term \textit{Remote-Clique} denotes the sum of pairwise distances in \citet{cevallosDiversityMaximizationDoubling2018}, but the mean of pairwise distances in \citet{coxDirectedDiversityLeveraging2021}.

\paragraph{Similar objectives, different realizations.}
Beyond name-computation collisions, several measures aim to capture the same concept but operationalize it differently. For example, both \citet{coxDirectedDiversityLeveraging2021} and \citet{yangMeasuringDataDiversity2025} partition the embedding space into discrete bins and apply Shannon entropy to bin counts. However, \citet{coxDirectedDiversityLeveraging2021} realize this via UMAP projection and 2D grid binning, and  \citet{yangMeasuringDataDiversity2025} via $K$-means clustering (``Partition Entropy'').

\subsection{Four Families of Diversity Measures}
\label{sec:taxonomy}
We survey existing measures (App.~\S\ref{app:taxonomy}) and organize them into four families (Tab.~\ref{tab:taxonomy}), based on the type of mathematical operation that transforms a set of embeddings into a scalar diversity score. We later use these families to make more general conclusions about the quality of different measurement approaches. The families are:

\paragraph{Distance-based} measures aggregate pairwise distances into a single score, without first creating additional structures such as clusters or graphs. An example is the frequently used Mean Pairwise Distance (e.g., in \citealt{tevetEvaluatingEvaluationDiversity2021, wang2025diversity, guoCuriousDeclineLinguistic2024}), which computes the mean cosine similarity across all embedding pairs.

\paragraph{Geometry-based} measures quantify the spatial properties of the embeddings. For example, the Convex Hull Volume measures the volume of the smallest convex region enclosing all embeddings \citep{yuCanDataDiversity2022} and Cluster Inertia is another popular example \cite{duBoostingDialogResponse2019,yangMeasuringDataDiversity2025, zhuMeasuringDiversitySynthetic2025}.

\paragraph{Distribution-based} measures quantify distributional properties in the embedding space, for example using entropy. %
An example measure is the Vendi Score, which is the exponential of the Shannon Entropy of the eigenvalues of the normalized similarity matrix \citep{friedman2022vendi}.

\paragraph{Graph-based} measures first build a graph with embedding vectors and then quantify its structural properties. For example, MST Dispersion is the total edge length of the minimum spanning tree of the fully connected graph over the embeddings, with pairwise distances as edge weights \citep{coxDirectedDiversityLeveraging2021}.\\

Some measures consist of steps that fall into different families. We resolve such cases based on their dominant step, or, where the dominant step is itself ambiguous, on the framing of the original work that introduced the measure. Cluster Inertia, for instance, partitions embeddings with $K$-means and then sums the squared distances from each embedding to its assigned centroid. We classify it as geometry-based (rather than distance-based) because it captures the spatial distribution of embeddings around cluster centroids.
 See App.~\S\ref{app:taxonomy} for a complete catalogue of measures with formulas, complexity, and references.

\newcolumntype{L}{>{\raggedright\arraybackslash}m{0.50\columnwidth}}
\newcolumntype{R}{>{\raggedright\arraybackslash}m{0.45\columnwidth}}

\begin{table}[t]
  \centering

  \footnotesize
  \setlength{\tabcolsep}{0pt}
  \begin{tabular}{@{}L @{\hspace{0.8em}} R@{}}
  \toprule
  \textbf{Distance-based}\newline
  {\itshape Aggregates pairwise distances; \newline no constructed structures.} &
  Mean Pairwise Dist.\newline Sum Pairwise Dist. \newline Energy \newline  Diameter\tablefootnote{Although the names of Diameter and Bottleneck suggest geometric constructs, they are simply the maximal and minimal pairwise distance, respectively; we therefore place them, together with their variants (Sum Diameter, Sum Bottleneck), in the distance-based family.}\newline Bottleneck\newline Sum Diameter\newline Sum Bottleneck \newline Chamfer Dist.\\
  \cmidrule(l{0pt}r{0pt}){1-2}
  \textbf{Geometry-based}\newline
  {\itshape Quantifies spatial properties of embeddings.} &
 Convex Hull Vol.\ (2D)\newline Span (Centroid)\newline Span (Medoid)\newline Cluster Inertia\newline Geo. Mean of Std. \newline Log-Determinant \\
  \cmidrule(l{0pt}r{0pt}){1-2}
  \textbf{Distribution-based}\newline
  {\itshape Quantifies distributional properties of embeddings.} &
   Vendi Score\newline R\'enyi Kernel Ent.\newline DCScore\newline Bins Entropy (UMAP)\newline MagArea \\
  \cmidrule(l{0pt}r{0pt}){1-2}
  \textbf{Graph-based}\newline
  {\itshape Summarizes a graph built over the embeddings.} &
  MST Dispersion\newline Graph Entropy\\
  \bottomrule
  \end{tabular}\\[3pt]
   \caption{Taxonomy of embedding-based diversity measures into four families. } \label{tab:taxonomy}
\end{table}

\section{Benchmark Design}
\label{sec:benchmark}
Our benchmark evaluates diversity measures along \textbf{three components of controlled variation} following \citet{stirlingGeneralFrameworkAnalysing2007}'s decomposition of diversity (\S\ref{subsec:axes})---variety, balance and disparity---instantiated across \textbf{two complementary tiers}:
a simulated data tier and a natural text tier (\S\ref{subsec:tiers}). 

\paragraph{Relational and controlled evaluation.} %
We evaluate measures relationally rather than absolutely, %
because in practice diversity measures are usually used comparatively (e.g., to choose among candidate corpora), which requires only that datasets be ordered correctly by diversity.
Concretely, for each component (see Fig.~\ref{fig:vbd-axes}) we build a sequence of five datasets---which we refer to as five diversity \emph{levels}---in which only that one component varies, while the others are fixed. 
We then check whether each measure responds in the correct direction across the sequence. Every (tier, component) configuration is replicated under 5 random seeds to separate stable measure behaviour from sampling variability.

\paragraph{Notation.} Three variables parameterize every dataset in the benchmark:
\begin{itemize}[itemsep=0.3em, parsep=0pt, topsep=1pt]
    \item $n$: the total number of elements in a dataset, fixed at $n = 10{,}000$ across all configurations.
    \item $k$: the number of \textbf{categories}
    in a dataset. A category is the basic diversity unit that an element belongs to. %
    \item $m$: the number of \textbf{super-categories} into which these $k$ categories are grouped, %
    with $m \leq k$. %
\end{itemize}
Each tier instantiates ``category'' and ``super-category'' differently: for the simulated data tier, categories are Gaussian ``centres'' and super-categories are ``super-centres'' that Gaussian centres are grouped under; %
for the natural text tier, categories are ``topics'' and super-categories are broader ``domains'' in Wikipedia's hierarchy.

\subsection{Three Components of Controlled Variation}
\label{subsec:axes}

The variety--balance--disparity decomposition originates in \citet{stirlingGeneralFrameworkAnalysing2007}, who unified definitions of diversity from ecology, economics, and science-and-technology studies into a single conceptual framework. Stirling argues that any substantive claim about diversity must distinguish three orthogonal questions (Fig.~\ref{fig:vbd-axes}): \emph{how many} categories are present (\textbf{variety}), \emph{how evenly} items are spread across those categories (\textbf{balance}), and \emph{how different} the categories themselves are from one another (\textbf{disparity}). Subsequent work has shown that each component can be operationalized independently, provided the data comes with a categorical structure \citep{vandamDiversityItsDecomposition2019}. In NLP, this concept of diversity was adopted as early as \citet{lion-bouton-etal-2022-evaluating}, but has been taken up more broadly only in the last few years \citep{esteveSurveyDiversityQuantification2025,nguyen-ploeger-2025-need}. %

\paragraph*{Mapping elements to categories.}
Applying the variety--balance--disparity decomposition requires deciding how to map \textit{elements} to \textit{categories}. This mapping can be performed in different ways. For example, it can be based on metadata in the dataset (e.g., explicit topic or language labels), inferred through clustering, or every element can be treated as its own category (e.g., a case discussed in \citealp{esteveSurveyDiversityQuantification2025}).
We note that many diversity dimensions of interest in NLP do not map cleanly onto categories. For example, although language varieties (e.g., dialects) are 
often treated as discrete entities, in practice the boundaries are less clear cut \citep{chambers1998dialectology}. See Limitations for a further discussion on this.

\paragraph*{Operationalization.} Our benchmark design adopts the variety--balance--disparity decomposition directly. Here, elements correspond to the vectors or texts in the considered dataset. 
In the simulated data tier, the categories are defined by the data-generating process. In the natural text tier, they are external metadata (Wikipedia topics) used as a proxy for semantic content. 
Note that the mapping is part of the benchmark's ground truth only; the measures themselves receive no category labels and only operate on the raw embeddings.

Each dataset in our benchmark contains $n=10{,}000$ elements that fall into a number of categories, which are themselves organized into a smaller number of super-categories. 
Then, \textit{variety} is the number of categories; \textit{balance} is how evenly the points are distributed across them; and \textit{disparity} is the separation between categories (Fig.~\ref{fig:vbd-axes}). 
Instead of manipulating the separation of categories directly, we use the number of super-categories as a proxy: fewer super-categories pack categories into related groups (low disparity), more super-categories spread them across unrelated ones (high disparity). 
The five diversity levels of each component are as follows.

\paragraph{Variety component.} A sequence of datasets with increasing number of categories $k \in \{10, 20, 30, 40, 50\}$, holding balance (i.e., uniform distribution) and disparity ($m=10$) fixed.

\paragraph{Balance component.} A sequence of datasets with five balance levels, ranging from uniform to strongly skewed, while holding variety ($k=50$) and disparity ($m=10$) fixed. The full set of conditions is summarized in App.~Tab.~\ref{tab:balance}.

\paragraph{Disparity component.} A sequence of datasets varying the number of super-categories $m \in \{10, 20, 30, 40, 50\}$, while holding the variety $k=50$ and balance (uniform distribution) fixed. For instance, at $m = 10$, the 50 categories cluster within 10 super-categories (low disparity); %
at $m = 50$, each category belongs to a distinct super-category (high disparity). %

\subsection{Two Complementary Tiers}
\label{subsec:tiers}

The simulated data tier (\S\ref{sec:simulated}) contains no text: data samples are drawn from Gaussian mixtures whose structure exactly realizes the three components, isolating measure behaviour from any embedding model effect. 
In contrast, the natural text tier (\S\ref{sec:wiki}) provides a more natural setting with factors that are more difficult to control. %
More details can be found in App.~\S\ref{app:benchmark}.

\subsubsection{Simulated Data Tier}
\label{sec:simulated}

Embedding-based diversity measures operate on a vector space in $\mathbb{R}^d$, so a clean test case is a simulated vector dataset where the three components of variation can be exactly controlled by construction parameters. The simulated data tier provides this: data is drawn from a hierarchical Gaussian mixture model. %
This tier functions as a %
test case with a clear ground truth: every factor is controlled by construction, category labels are not noisy proxies but are part of the data-generating structure itself, and there is no embedding model in the loop. 
The flip side of this realization is that simulated  samples of vector spaces are isotropic and Gaussian, while real text embeddings occupy a more structured and non-Gaussian space \citep{caiIsotropyContextualEmbedding2020, timkeyAllBarkNo2021}; the following tier introduces this complexity from real text.

\paragraph{Construction conditioned on $n, k$ and $m$.} We sample $m$ super-category centres $\mathbf{s}_1,\dots,\mathbf{s}_m$ from a Gaussian %
$\mathcal{N}(\mathbf{0}, \sigma_{\text{super}}^2 I)$. Next, we sample $k$ centres. For that, we first distribute each category as evenly as possible over the $m$ super-centres. For each of the $k$ categories, we then draw a category centre $\mathbf{c}_i\sim\mathcal{N}(s_{m(i)},\,\sigma_{\text{category}}^2 I)$, where $m(i)$ denotes the super-centre assigned to category i. Then, each element of category $i$ is drawn from $\mathcal{N}(\mathbf{c}_i,\,\sigma_{\text{element}}^2 I)$. The three scale parameters satisfy $\sigma_{\text{super}} > \sigma_{\text{category}} > \sigma_{\text{element}}$, so super-centres are well separated, category centres are grouped within their super-centre, and data points are grouped within their centre. Details can be found in App.~\S\ref{app:gmm}.

As a result of the simulated construction, each component of diversity can be manipulated independently: the number of category centres $k$ controls variety; the distribution of the fixed $n$ data points across the $k$ categories controls balance; and the number of super-centres $m$ controls disparity---
because super-centres are far apart while category centres cluster tightly around their assigned super-centre, a small $m$ bunches the $k$ categories into a few tight groups (low disparity), while a large $m$ spreads them apart (high disparity).

\subsubsection{Natural Text Tier}
\label{sec:wiki}
Next, we look at a natural text setting. %
We use semantic diversity as an example, because this is the diversity dimension that embedding-based measures have primarily been applied to and it is one of the main dimensions of diversity that researchers have been interested in
\cite{tevetEvaluatingEvaluationDiversity2021, hanMeasuringImprovingSemantic2022}.
This tier uses real natural language text drawn from Wikipedia, making it more directly relevant to downstream use. %

Annotating semantics directly at the sentence level is difficult and often ambiguous \cite{pavlickInherentDisagreementsHuman2019}, so we operationalize it through \emph{topic-level categorical structure}: a sentence's category is the Wikipedia topic of the article it is drawn from, and the topic serves as a proxy for semantic content. Concretely here, domains correspond to the broader super-centres and topics to smaller centres in the simulated data tier; App.~Tab.~\ref{tab:hierarchy} illustrates its structure with examples. 

\paragraph{Construction.} The construction has two stages: selecting domains (super-categories) and topics (categories), and collecting sentences from them.

(1) To obtain domains that are semantically well-spread rather than clustered in a single region of the Wikipedia hierarchy, we embed every domain label with embedding model Qwen3-Embedding-8B\footnote{\href{https://huggingface.co/Qwen/Qwen3-Embedding-8B}{\nolinkurl{Qwen/Qwen3-Embedding-8B}}} and select via \textbf{farthest-point sampling} (FPS), a standard spread-maximizing algorithm that iteratively picks the candidate maximally distant from those already chosen. Within each selected domain, we apply FPS again over its topics. %

(2) For each selected topic, we then collect its sentences from the category tree. First, we draw from articles directly under the topic, descending into subcategories only when there are not enough sentences; this avoids the semantic drift that can occur when digging too far down the category tree. Second, we cap each article's contribution at five sentences (which prevents any single article from dominating a topic) and sample them randomly %
(which avoids overrepresenting the stylized leading sentences characteristic of Wikipedia articles). Further details are in App.~\S\ref{app:wiki-pool}.

\paragraph{Label noise.} Because sentences in our dataset are sampled from articles under a topic's category, not every sentence is necessarily about its Wikipedia topic, and we treat such cases as noise. For example, the sentence ``Gerhard Fisher studied electronics at the University of Dresden before emigrating to the United States.'' is a background sentence about the inventor of the hand-held metal detector, a device used in prisons, but the sentence itself is not semantically related to its Wikipedia topic \emph{Prisons}. Noise like this comes with using real-world text data, so instead of removing it, we quantify it: one of the authors samples 750 sentences from all 75 datasets (i.e., 10 sentences per dataset) and checks whether each sentence is semantically about its Wikipedia topic. 
About 17.9\% of the sentences do not support their Wikipedia topics. Within each component, the noise rate does not vary substantially across the five levels, so it degrades all five datasets alike, and also the levels are separated by wide intervals ($k$ and $m$ both move in steps of 10), so the noise should not affect the relative ordering of datasets by component. We therefore do not expect label noise to invalidate the benchmark. More details can be found in App.~\S\ref{app:noise}.
\section{Results}
\label{sec:results}

We display results for both the simulated data and natural text tier in Fig.~\ref{fig:results-d384-mean}. %
For the diversity components, we denote V for variety, B for balance and D for disparity.
\input{figures/fig_synth-wiki_combined}

\label{sec:protocol}

\paragraph{Evaluation metric.} For each diversity component (\S\ref{subsec:axes}), %
each tier (simulated data and natural text; \S\ref{subsec:tiers}), and each seed (5 seeds), we obtain a sequence of 5 datasets that systematically vary along one single component. %
For each diversity measure $M$, we compute $M$ on every dataset in the sequence and check whether its values induce the correct dataset ordering with respect to the component being varied.
We report Spearman's rank correlation $\rho$ between the ground truth ordering and values of $M$, averaged over five seeds. %

\paragraph{Embedding configurations.} For the natural text tier, we use embedding model MiniLM-L6-v2\footnote{\href{https://huggingface.co/sentence-transformers/all-MiniLM-L6-v2}{\nolinkurl{sentence-transformers/all-MiniLM-L6-v2}}} ($d=384$) \cite{reimersSentenceBERTSentenceEmbeddings2019}, since the model is optimized %
to align semantically similar sentence pairs. The simulated data tier uses no embedding model, but instead is sampled from a $d=384$-dimensional vector space. %

\subsection{Analysis and Discussion}

\paragraph{No measure responds to a single component only.} We regard a measure as \emph{component-specific} if it responds to a single component ($\rho \ge 0.9$) while remaining insensitive to the other two ($|\rho| \le 0.2$). By this criterion, no measure is component-specific on either tier: those responding strongly to one component always respond to at least one other (Fig.~\ref{fig:results-simulated-d384-mean}, \ref{fig:results-natural-text-d384-mean}).

\paragraph{Mean Pairwise Distance (a frequently used measure) underperforms.}
Mean Pairwise Distance is a common choice in prior work \citep{tevetEvaluatingEvaluationDiversity2021,shaibStandardizingMeasurementText2025,mirandaScaleDiversityCoefficient2025}. It performs well on our simulated data; however, even on this easy and controlled setting, it does not perform perfectly (mean $\rho < 1$ on both variety and balance; Fig.~\ref{fig:results-simulated-d384-mean}). On natural text, the situation is worse: on average, its correlation drops to roughly half its value on the simulated data tier (Fig.~\ref{fig:results-natural-text-d384-mean}). On both tiers, several alternatives---Vendi $q{=}1.0$, R\'enyi Kernel Entropy, and Graph Entropy---consistently outperform Mean Pairwise Distance across the three components.

\paragraph{Real text is more challenging than the simulated data setting.}
On the simulated data tier (Fig.~\ref{fig:results-simulated-d384-mean}), measures generally perform well, but a few perform ``catastrophically'' (high negative correlation) on some components which leads to a lower mean than median. %
Six distinct measures reach a mean $\rho \ge 0.9$ and across-seed standard deviation $\le 0.1$ for simulated data on all three components. 
In contrast, on natural text (Fig.~\ref{fig:results-natural-text-d384-mean}), no measure reaches this bar and the gap between mean and median closes. The weakening is no longer driven by a few catastrophic measures but is broad, with nearly all measures affected. Some of the top measures on the simulated tier drop severely (e.g., Log-Determinant drops from $+1.00$ to $+0.24$ on disparity in Fig.~\ref{fig:results-d384-mean}).
The difficulty of the natural text tier might come from the noise of real text (App.~\S\ref{app:noise}) or from properties of the embedding space produced by the used embedding models, or from both. The drop %
demonstrates the necessity to benchmark measures on real text as well.  %

\paragraph{Disparity is the hardest component on both tiers.}
For both tiers, disparity reaches the lowest mean and median scores. 
For the simulated data tier (Fig.~\ref{fig:results-simulated-d384-mean}), variety scores are higher, but for natural text (Fig.~\ref{fig:results-natural-text-d384-mean}), variety scores are lower than balance scores. %

\paragraph{On both tiers, every family contains comparatively good measures and the best/worst ones stay the same.} Averaged over the three components, graph- and distribution-based measures are generally stronger than geometry- and distance-based ones on both tiers, %
and graph-based measures degrade least across tiers ($0.10$, versus $0.35$ for distribution). Even with the noticeable gap between family averages, each family contains a strong member: the six measures meeting our bar (mean $\rho \ge 0.9$ with SD $\le 0.1$) on the simulated tier span all four families, and on natural text the best member of each family lies within a gap of $0.11$ (Chamfer Dist.\ $+0.82$, Cluster Inertia $+0.85$, Vendi $+0.77$, Graph Entropy $+0.88$). Therefore, we do not regard any family as inherently inferior, as a poorly designed measure can arise in any family. So a low average is evidence about the members we survey rather than about the approach itself. Across tiers, Graph Entropy performs the best and Convex Hull Volume (CHV) the worst. 

\section{Robustness Checks}
Sentence embeddings are known to be sensitive to model choice and their dimensionality \cite{ethayarajhHowContextualAre2019, tsukagoshiRedundancyIsotropyIntrinsic2025}. Because our results are produced on top of one fixed embedding setup, we test the robustness of our findings along two aspects: 
On the simulated data tier we vary the dimension $d$ of the GMM (\S\ref{sec:robust-dim}); on the natural text tier we vary the embedding model used to encode the sentences (\S\ref{sec:robust-model}).

\subsection{Dimensionality}\label{sec:robust-dim}
We rerun the simulated data (GMM) tier at five dimensions $d \in \{256, 384, 1024, 2048, 4096\}$. %
Results are shown in App.~Fig.~\ref{fig:cross-dimensions}.

\paragraph{Most findings are stable across dimensions, with a few exceptions.}
Four of the six measures that perform well (i.e., $\rho \geq 0.9$ with SD $\leq 0.1$ on all three components) at $d=384$---Vendi ($q{=}1$), Rényi Kernel Entropy, DCScore and Graph Entropy---also do so for all other tested dimensions; Mean Pairwise Distance falls below the bar at $d=256$ and $d=1024$, and Log-Determinant at $d=2048$.
At the family level, the distribution- and graph-based measures stay the strongest, and the distance- and geometry-based measures the weakest. 
The distance family, however, overtakes geometry at every dimensionality other than $d=384$, though their mean $\rho$ differs by at most $0.13$ at any dimensionality; we therefore do not read their relative order as a stable finding.

\subsection{Embedding models}\label{sec:robust-model}
We re-encode the natural text tier datasets with two different semantic embedding models (\emph{Qwen3-Embedding-8B\footnote{\href{https://huggingface.co/Qwen/Qwen3-Embedding-8B}{\nolinkurl{Qwen/Qwen3-Embedding-8B}}}} and \emph{e5-mistral-7b-instruct\footnote{\href{https://huggingface.co/intfloat/e5-mistral-7b-instruct}{\nolinkurl{intfloat/e5-mistral-7b-instruct}}}}) and an embedding model sensitive to style rather than semantics (\emph{Style Embedding\footnote{\href{https://huggingface.co/AnnaWegmann/Style-Embedding}{\nolinkurl{AnnaWegmann/Style-Embedding}}}} from \citealp{wegmann-etal-2022-author}). %
Results are shown in App.~Fig.~\ref{fig:cross-embedders} and Fig.~\ref{fig:cross-embedders-style}.

\paragraph{Embeddings need to capture the intended diversity dimension.} As expected, diversity measures using Style Embedding cannot capture %
semantic diversity.

\paragraph{Findings can differ for a different semantic embedding.}
Most findings with MiniLM hold for Qwen3 and E5 (App.~\S\ref{app:dim-robustness}), but individual measures can diverge. For example, Mean Pairwise Distance is weak on both balance and variety with MiniLM, reasonable on both with Qwen3, and strong on balance but weak on variety with the similarly-sized E5.\footnote{Because Qwen3 and E5 are optimized for retrieval and ranking rather than for embedding distances that reflect semantic similarity, %
we base our main conclusions on MiniLM.} We leave further investigation to future work.

\section{Conclusion}
\label{sec:conclusion}

In this work, we collect and categorize 21 embedding-based diversity measures into four families (Tab.~\ref{tab:taxonomy}). %
Further, we introduce the \mbox{\textsc{EmbDivBench}} benchmark to evaluate diversity measures along three components: variety, balance and disparity. We evaluate the measures in a perfectly controlled simulated data setting---independent of a specific definition of diversity (e.g., stylistic vs. semantic). Surprisingly, we find that a majority of measures already behave unexpectedly on this easy and controlled setting. Additionally, \textsc{EmbDivBench} includes a semantic tier to exemplify a specific diversity dimension (i.e., semantic) with real text data. The real data setting shows that the requirements for diversity measures might be quite different on an embedding space as more measures fall short of our expectations. No measure performs perfectly. 
We invite the community to expand our benchmark with other real data settings for different diversity dimensions. All measures benchmarked here and more are implemented in the \texttt{emb-diversity}\footnote{\url{https://github.com/nlpsoc/emb-diversity/}} package, which we release as follow-up work \citep{su2026embdiversitytoolembeddingbasedmeasurement}.  The package makes it straightforward to apply measures to new data, or to add newly proposed measures and compare them against the ones surveyed here. %

\paragraph{Recommendations.}
The most suitable diversity measure depends on various factors, including the diversity component of interest, the embedding model chosen, and the available computational budget (see App.~Tab.~\ref{tab:cost} for computational cost of each measure). 
We make the following recommendations based on our empirical benchmark results:\label{recommendations}
\begin{itemize}[itemsep=0.3em, parsep=0pt, topsep=1pt]
    \item \textbf{Do not use Mean Pairwise Distance.} Although intuitive and popular, Mean Pairwise Distance performs worse than several other measures on simulated and text data.
    \item \textbf{Select measures based on the component(s) of interest.} 
    No measure performs best on all three components. We therefore recommend identifying the component of interest first and consulting our %
    component-specific shortlists. Our recommendations should be read as the currently best available option rather than a future-proof quality judgement.
    \begin{itemize}
        \item \textbf{Variety:} Chamfer Distance\footnote{Sum Bottleneck is rank-equivalent to Chamfer Distance at a fixed dataset size and therefore interchangeable here (App.~\S\ref{app:taxonomy}).}%
        and MST Dispersion %
        \item \textbf{Balance:} Graph Entropy
        \item \textbf{Disparity:} 
        MST Dispersion %
    \end{itemize}

    If no single component is of primary interest, Graph Entropy is currently the best-supported single choice, followed by Vendi ($q{=}1$). See App.~\S\ref{app:recommendations} for the detailed rationale.

    \item \textbf{Choose your embedding model carefully.} Choose an embedding model that (primarily) captures the dimension of diversity you want to capture (e.g., semantic diversity). %
    Even among models targeting the same diversity dimension, differences between embedding models can affect measure behaviour and rankings (\S\ref{sec:robust-model}). We recommend practitioners to confirm that the embedding they use is sensitive and faithful to their diversity dimension of interest. %
\end{itemize}

\section*{Limitations}

\paragraph{Categorical structure required.} The V-B-D decomposition needs a partition of the data into categories: variety counts them, balance is the distribution over them, and disparity is the separation between them. Both our tiers provide one by construction, but some diversity dimensions are graded rather than categorical (e.g., formality is relative, see \citealp{wegmann_survey_2026}). 
This raises a broader question for future work: to what extent the V--B--D decomposition, which was developed outside NLP, is suitable for measuring diversity in natural language data. %
Future work could also investigate alternative mappings from texts to categories: some measures already carry an implicit mapping (e.g., DCScore treats each sample as its own category; App.~\S\ref{app:taxonomy}), and a measure whose implicit mapping aligns with the categories used to construct a dataset may be favoured by that alignment.

\paragraph{Confounds inherited from Wikipedia.}
First, topic labels are an imperfect proxy: not all sentences within a given article are directly related to the article’s topic, introducing noise into the topic clusters (see our analysis in App.~\S\ref{app:noise}).
Second, disparity on our natural text tier is operationalized as the number of domains spanned. This proxy assumes that cross-domain topics lie further apart in embedding space than within-domain topics, but Wikipedia's editorial taxonomy does not enforce this. For example, the topic \textit{History of mathematics} (under domain History) and topic \textit{Mathematician} (under domain Mathematics) belong to different domains yet are semantically close. 

\paragraph{Our benchmark only captures part of the picture.}
Our benchmark tests the measures against the three diversity components and only considers the ordering of datasets; it does not consider the raw diversity scores or change in those scores. 
Future work could investigate to what extent score differences are perceptible to humans or have meaningful downstream effects on trained models \citep{kocmi-etal-2024-navigating}. Our work also does not test the formal axiomatic properties (monotonicity, uniqueness, continuity) studied by \citet{mironov2024measuring}; these two approaches are complementary, as a measure may satisfy these axioms yet respond weakly on our three-component benchmark. Other aspects, such as actionability \citep{delobelle_metrics_2024}, should also inform measure selection.

\paragraph{Hyperparameter settings untested.} Several measures involve hyperparameters that we fix to a single value---typically a published default or a setting inherited from prior work 
(see App.~Tab.~\ref{tab:hyperparams}). A systematic sweep over these parameter values may shift the relative rankings we report.

\paragraph{Sentence-level units and English Wikipedia only.} Our benchmark does not evaluate the measures on paragraph- or document-level units, which would require different pipelines (mean-pooling, sliding-window aggregation, or long-context encoders).
Whether the relative behaviour of measures we report transfers to diversity measurement with longer units remains therefore an open question for future work.
Also, our natural text tier is built exclusively from English Wikipedia, whose category hierarchy, article-length distribution, and topical coverage reflect the editorial choices of one language community.
Still, the taxonomy and pipeline we introduce are not language-specific, and could be extended to other Wikipedia languages in future work.

\paragraph{Embedding model artefacts.} Because the measurements in the natural text tier are computed over sentence embeddings, they are also influenced by properties of the embedding space itself (anisotropy, hubness) in addition to properties of the underlying data. %
One concrete step for future work would be calibrating the simulated tier to real embedding statistics: e.g., drawing simulated data from distributions whose covariance is estimated from real sentence embeddings instead of isotropic Gaussians, to better simulate the anisotropy of natural embedding spaces, which %
would help clarify the extent to which the performance drop from the simulated to the natural text tier (\S\ref{sec:results}) can be explained by embedding-space artefacts.

\section*{Ethical Considerations}

\paragraph{Risk of misuse, PII and offensive content.}
Our work is methodological and introduces no new data or models: all text is from English Wikipedia, and the released benchmark consists of dataset configurations and evaluation code only. We apply no name redaction, as public-figure names in biographical articles are public information and carry no contact or identification details; we also apply no offensive-content filtering, as sensitive topics (e.g., controversial figures) appear in encyclopaedic context. We do not foresee ethical risks beyond standard methodological caveats.

\section*{Acknowledgments}
This work is supported by the ERC Starting Grant DataDivers 101162980.

\begin{CJK*}{UTF8}{gbsn}
\bibliography{references_cleaned}
\end{CJK*}

\appendix
\label{sec:appendix}

\section{Detailed Taxonomy of Embedding-based Diversity Measures} \label{app:taxonomy}
\definecolor{distcol}{HTML}{2C6FB5}   %
\definecolor{geomcol}{HTML}{2E8B6E}   %
\definecolor{graphcol}{HTML}{D08A1D}  %
\definecolor{distrcol}{HTML}{7E57C2}  %

\begin{figure}[ht]
  \centering
  \resizebox{\columnwidth}{!}{%
  \begin{tikzpicture}[>={Latex}]

    \draw[gray!55, line width=0.8pt, -{Latex[length=5pt]}] (1.2,1.2)--(8.8,1.2);
    \draw[gray!55, line width=0.8pt, -{Latex[length=5pt]}] (1.2,1.2)--(1.2,6.0);
    \node[gray!60, font=\scriptsize\itshape, anchor=south east]
          at (8.78,1.27) {embedding space $\mathbb{R}^{d}$};

    \draw[geomcol, line width=1pt, fill=geomcol!8]
      plot[smooth cycle, tension=0.75]
      coordinates {(1.7,3.3) (2.2,4.25) (3.3,4.62) (4.6,4.2)
                   (5.1,3.2) (4.7,2.2) (3.4,1.8) (2.1,2.2)};

    \foreach \tx/\ty in {2.0/2.8, 3.0/2.3, 3.9/2.4, 4.5/2.9, 3.1/4.0}{
      \draw[graphcol, line width=0.9pt, dashed, -{Latex[length=3.5pt]}]
            (3.4,3.1)--(\tx,\ty);}

    \draw[distcol, line width=1.1pt,
          {Latex[length=4pt]}-{Latex[length=4pt]}] (2.3,3.6)--(4.2,3.7);
    \node[distcol, font=\scriptsize] at (3.0,3.34) {$d_{ij}$};

    \foreach \px/\py in {2.3/3.6, 3.1/4.0, 2.0/2.8, 4.2/3.7,
                         4.5/2.9, 3.0/2.3, 3.9/2.4}{
      \fill[black!75] (\px,\py) circle (2pt);}
    \fill[black!80] (3.4,3.1) circle (2.6pt);

    \draw[distrcol, line width=0.9pt, -{Latex[length=5pt]}]
          (4.9,4.05) to[out=35,in=170] (6.95,5.0);
    \draw[gray!55, line width=0.6pt, -{Latex[length=3.5pt]}] (7.0,4.5)--(8.4,4.5);
    \draw[gray!55, line width=0.6pt, -{Latex[length=3.5pt]}] (7.0,4.5)--(7.0,5.55);
    \foreach \sx in {7.3,7.45,7.6,7.75,7.9}{
      \draw[distrcol!55, line width=0.7pt] (\sx,4.5)--(\sx,4.62);}
    \draw[distrcol, line width=1.1pt]
      plot[smooth, tension=0.85]
      coordinates {(7.05,4.56) (7.35,4.8) (7.62,5.28) (7.92,4.82) (8.28,4.56)};

    \node[distcol, font=\small\bfseries, fill=white, inner sep=1.6pt]
          (dl) at (3.0,5.0) {Distance-based};
    \draw[distcol, line width=0.5pt] (dl.south)--(2.65,3.63);

    \node[graphcol, font=\small\bfseries, fill=white, inner sep=1.6pt]
          (gl) at (6.15,2.3) {Graph-based};
    \draw[graphcol, line width=0.5pt] (gl.west)--(4.05,2.72);

    \node[geomcol, font=\small\bfseries, anchor=west] (yl) at (5.95,4.0)
          {Geometry-based};
    \draw[geomcol, line width=0.6pt] (yl.west)--(5.02,3.6);

    \node[distrcol, font=\small\bfseries] at (7.66,5.78) {Distribution-based};

  \end{tikzpicture}}
  \caption{Illustration behind the four measure families on a toy vector space.} %
  \label{fig:family-intuition}
\end{figure}
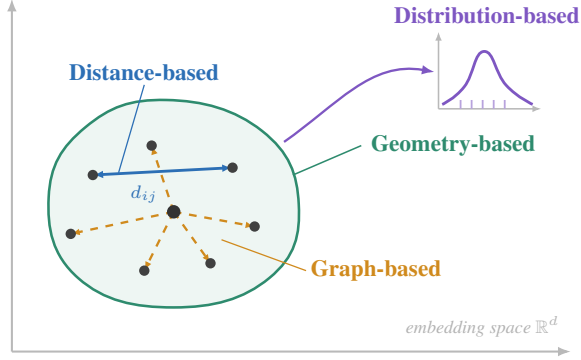

This Appendix presents the full taxonomy of embedding-based diversity measures surveyed in this work, organized by their underlying mathematical formulation. Fig.~\ref{fig:family-intuition} provides a visual 
intuition for the four families.

\paragraph*{Notation and Conventions}

Throughout this Appendix, we use the following notation:
\begin{itemize}
    \item $X = \{x_1, \ldots, x_n\}$: a dataset of $n$ samples
    \item $\mathbf{e}_i \in \mathbb{R}^d$: the $d$-dimensional embedding of sample $x_i$
    \item $d_{ij} = d(\mathbf{e}_i, \mathbf{e}_j)$: distance between embeddings (cosine distance, unless specified)
    \item $s_{ij} = sim(\mathbf{e}_i, \mathbf{e}_j)$: similarity between embeddings (cosine similarity, unless specified)
\end{itemize}

\paragraph{Comparability across dataset sizes.} Several measures are defined as sums over elements or pairs (e.g.\ Sum Pairwise Distance, Sum Diameter, Sum Bottleneck, Graph Entropy, MST Dispersion) and therefore grow with $n$, so their values are not comparable across datasets of different sizes. Since $n$ is fixed at 10{,}000 across every dataset in our benchmark, this does not affect any result we report.

\subsection{Distance-based Measures}

This family directly measures diversity using raw pairwise distances or similarities, without building any additional structures such as graphs, cluster assignments, or geometric constructs (e.g., convex hulls) that require further processing beyond distance computation.

\subsubsection{Mean Pairwise Distance}
\noindent\textbf{Intuition.}
This measure averages the distance of each sample to all other samples in the embedding space. 
A higher value indicates that samples are further apart (i.e., more diverse), whereas a lower value implies less diversity.

\noindent\textbf{Computation.}
Compute the average of all off-diagonal pairwise distances:
\begin{equation}
AvgDist(\mathbf{X})
= \frac{2}{n(n-1)} \sum_{i<j} d_{ij}
\end{equation}
In principle, this measure generalizes to any function capable of quantifying similarity or distance between two samples. However, within the context of embeddings, \textbf{cosine similarity} is the most prevalent choice. When employed, this measure is often referred to as \textbf{Remote-Clique} \citep{coxDirectedDiversityLeveraging2021}. %
Some works also refer to the similarity-based variant \textbf{Self-cosSim} \citep{tevetEvaluatingEvaluationDiversity2021,  zhang-etal-2024-improving-diversity, mirandaScaleDiversityCoefficient2025}. Since cosine similarity is $1 - \text{cosine distance}$, these measures are mathematically equivalent (linearly related).

\noindent\textbf{Complexity.}
$\mathcal{O}(n^2 d)$
\\

\subsubsection{Sum Pairwise Distance}
\noindent\textbf{Intuition.}
By summing all pairwise distances, this measure captures the overall spread of samples in the embedding space. Different terms refer to the same formulation, originally introduced as Max Dispersion (MD) by \citet{yuCanDataDiversity2022}
and later adopted as DistSum by \citet{yangMeasuringDataDiversity2025}.\\
Higher values indicate greater diversity.

\noindent\textbf{Computation.}
Compute the sum of all pairwise distances:
\begin{equation}
\mathrm{MD}(\mathbf{X}) = \mathrm{DistSum}(\mathbf{X}) 
= \sum_{i<j} d_{ij},
\end{equation}

\noindent\textbf{Complexity.}
$\mathcal{O}(n^2 d)$
\\

\subsubsection{Chamfer Distance}
\noindent\textbf{Intuition.}
This measure originates from computational geometry and computer vision (specifically point cloud processing), where it is used to compare \emph{two} sets of points by measuring how close every point in one set is to its nearest neighbour in the other set.
For point sets $A$ and $B$, the Chamfer Distance averages: 
(i) the distance from each point in $A$ to its nearest neighbour in $B$, and 
(ii) the distance from each point in $B$ to its nearest neighbour in $A$.

In the context of diversity research in NLP, prior work adapts this idea to a \emph{single} dataset \citep{coxDirectedDiversityLeveraging2021, zhang2025evaluating}. 
Instead of comparing two sets, the measure averages, for each sample, the distance to its nearest neighbour. This reflects the \emph{local compactness} of the embedding space: datasets with tighter clusters have smaller nearest-neighbour distances, while more dispersed datasets yield larger values.

\noindent\textbf{Computation.}
Compute the average of the minimum pairwise distances:
\begin{equation}
\mathrm{Chamfer}(\mathbf{X})
= \frac{1}{n} \sum_{i=1}^{n} \min_{j \neq i} d_{ij}
\end{equation}

\noindent\textbf{Complexity.}
$\mathcal{O}(n^2 d)$
\\

\subsubsection[Energy(gamma)]{Energy$(\gamma)$}
\noindent\textbf{Intuition.}
This measure originates from classical mechanics. It takes a dataset as a system of equally charged particles distributed in the embedding space and measures its diversity as the negative potential energy \citep{velikonivtsevChallengesGeneratingStructurally2024,mironov2024measuring}.  
Each pair of samples repels each other, inversely proportional to their distance raised to the power $\gamma$.  
The measure encourages uniform spacing and penalizes close pairs; for exact duplicates ($d_{ij}=0$), the measure diverges to $-\infty$.
Thus, when samples are further apart, the negative potential energy (and hence the diversity score) increases.  

\noindent\textbf{Computation.}
The Energy measure of parameter $\gamma > 0$ is defined as:
\begin{equation}
\mathrm{Energy}_\gamma(\mathbf{X})
= -\,\frac{2}{n(n-1)} 
  \sum_{i<j} \frac{1}{d_{ij}^{\,\gamma}}.
\end{equation}
The negative sign ensures that configurations with smaller pairwise distances
(i.e.\ higher repulsive energy) receive lower diversity values,
so that increasing all pairwise distances increases the overall diversity score.  
The value is always negative, and higher values (i.e. closer to zero) indicate greater diversity.
In the special case of $\gamma = 1$, the measure corresponds to the potential energy of a system of point charges under Coulomb repulsion.

\noindent\textbf{Complexity.}
$\mathcal{O}(n^2 d)$ (pairwise distance computation for all sample pairs)
\\

\subsubsection{Diameter (maximal pairwise distance)}
\noindent\textbf{Intuition.}
This measure finds the largest distance between any two embeddings  of a dataset \citep{mironov2024measuring, xieHowMuchSpace2023}.  
It shows how far apart the most different embeddings are in the space, figuratively the “width” of the point cloud.

\noindent\textbf{Computation.}
The Diameter is the largest of these pairwise distances, defined as:
\begin{equation}
\mathrm{Diameter}(\mathbf{X})
= \max_{i<j} d_{ij}
= \max_{i<j} d(\emb{i}, \emb{j}).
\end{equation}

\noindent\textbf{Complexity.}
$\mathcal{O}(n^2 d)$ (pairwise distance computation)\\

\subsubsection{Bottleneck (minimal pairwise distance)}
\noindent\textbf{Intuition.}
This measure computes  the smallest distance between any two data points  \citep{mironov2024measuring, xieHowMuchSpace2023}.  

\noindent\textbf{Computation.}
The Bottleneck is defined as:
\begin{equation}
\mathrm{Bottleneck}(\mathbf{X})
= \min_{i<j} d_{ij}.
\end{equation}

\noindent\textbf{Caveats.}
As a stand-alone diversity measure, this measure can be misleading---a single pair of near-duplicates can collapse its value regardless of the rest of the distribution.

\noindent\textbf{Complexity.}
$\mathcal{O}(n^2 d)$\\

\subsubsection{Sum Diameter (sum of maximal pairwise distances)}
\noindent\textbf{Intuition.}
This measure sums up the maximum distance from each data point to its farthest data point. 
While the Diameter measure relies on a single pair of extreme points (making it sensitive to outliers), Sum Diameter aggregates the maximum separation for every data point \citep{mironov2024measuring, xieHowMuchSpace2023}.  

\noindent\textbf{Computation.}
For each sample $x_i$, find its maximum distance to all other samples and sum these maxima:
\begin{equation}
\mathrm{SumDiameter}(\mathbf{X})
= \sum_{i=1}^{n} \max_{j \ne i} d_{ij}.
\end{equation}

\noindent\textbf{Complexity.}
$\mathcal{O}(n^2 d)$\\

\subsubsection{Sum Bottleneck (sum of minimal pairwise distances)}
\noindent\textbf{Intuition.}
While the Bottleneck measure is determined only by the closest pair of samples (making it sensitive to (near) duplicates), this measure aggregates the minimum distance from each data point to its nearest neighbour for every data point \citep{mironov2024measuring, xieHowMuchSpace2023}. %

\noindent\textbf{Computation.}
For each sample $x_i$, find its minimum distance to all other samples and sum these minima:
\begin{equation}
\mathrm{SumBottleneck}(\mathbf{X})
= \sum_{i=1}^{n} \min_{j \ne i} d_{ij}.
\end{equation}

\noindent\textbf{Caveats.}
Unlike Diameter or Sum Diameter, this measure primarily reflects \emph{local} density rather than the overall spread of the dataset. %

\noindent\textbf{Complexity.}
$\mathcal{O}(n^2 d)$\\

\subsection{Geometry-based Measures}
This family treats the embeddings of a dataset as a point cloud in $\mathbb{R}^d$ and quantifies diversity through the geometric properties of that configuration---its extent, volume, or spread around a representative point, etc. %

\subsubsection{Convex Hull Volume}\footnote{In our benchmark we compute CHV after reducing embeddings to two dimensions with UMAP, as direct computation in the original embedding space ($d=384$) is intractable.}
\noindent\textbf{Intuition.}
This measure treats each embedding as a data point. The convex hull then represents the smallest convex region containing all points in the embedding space \citep{yuCanDataDiversity2022}.
A higher value indicates that the samples are more widely distributed, indicating greater diversity.

\noindent\textbf{Computation.}
The convex hull $\mathcal{H}(\mathbf{X})$ is defined as the smallest convex set containing all embeddings:
\[
\mathcal{H}(\mathbf{X}) = 
\left\{ \sum_{i=1}^{n} \alpha_i \emb{i} \,\middle|\, 
\alpha_i \ge 0, \, \sum_{i=1}^{n} \alpha_i = 1 \right\}.
\]
The Convex Hull Volume is then the $d$-dimensional volume of $\mathcal{H}(\mathbf{X})$:
\begin{equation}
\mathrm{CHV}(\mathbf{X}) = 
\mathrm{Vol}_d \big( \mathcal{H}(\mathbf{X}) \big).
\end{equation}
In practice, the volume can be efficiently approximated using the Quickhull algorithm \citep{barber1996quickhull}, 
which recursively partitions the space to compute the convex envelope of the points.  
When embeddings are high-dimensional, the algorithm operates on the subset of boundary points that define the hull.

\noindent\textbf{Complexity.}
Expected $\mathcal{O}(n \log n)$ using Quickhull

\subsubsection{Geometric Mean of Standard Deviations}\label{app:tax:geomeanstd}
\noindent\textbf{Intuition.}
This measure was introduced by \citet{laiDiversityDensityHomogeneity2020} under the name Diversity and adopted by \citet{yangMeasuringDataDiversity2025} under the name Radius. It is the geometric mean of the per-dimension standard deviations of the embeddings. Treating each embedding dimension as an independent axis of variation, 
the per-dimension standard deviations define an axis-aligned ellipsoid whose semi-axis lengths are $\sigma_1, \ldots, \sigma_d$. The product $\prod_j \sigma_j$ is proportional to the volume of this ellipsoid. 
Taking its $d$-th root converts this product into an average per-axis spread.
A larger value indicates greater dispersion, and hence greater diversity.

\noindent\textbf{Computation.}
Let $\sigma_j$ denote the standard deviation of the $j$-th embedding dimension across all samples. The measure is defined as the geometric mean of these per-dimension standard deviations:
\begin{equation}
\mathrm{GeoMeanStd}(\mathbf{X}) = \left( \prod_{j=1}^{d} \sigma_j \right)^{\!\frac{1}{d}}.
\end{equation}
Note that the measure collapses to zero whenever any standard deviation is zero.

\noindent\textbf{Complexity.}
$\mathcal{O}(n d)$ (computing the standard deviation along each embedding dimension).

\subsubsection{Span (Centroid)}
\label{app:tax:spancentroid}
\noindent\textbf{Intuition.}
This measure computes the average distance from each embedding to the centroid of all embeddings in the dataset \citep{coxDirectedDiversityLeveraging2021}.  
The centroid represents the centre of the embedded dataset, and Span quantifies how far, on average, the embeddings deviate from this central point. 
Datasets with larger Span values are more widely distributed around the mean embedding, indicating higher diversity.

\noindent\textbf{Computation.}
First, compute the centroid of all embeddings:
\begin{equation}
\boldsymbol{\mu} = \frac{1}{n} \sum_{i=1}^{n} \emb{i}.
\end{equation}
Then, compute the set of distances from all embeddings to this centroid:
\[
D = \{ d(\emb{i}, \boldsymbol{\mu}) \mid i = 1, \ldots, n \}.
\]
To ensure robustness against outliers, \citet{coxDirectedDiversityLeveraging2021} define the measure not as the maximum, but as the \textbf{90th percentile} of these distances:
\begin{equation}
\mathrm{Span}_{\mathrm{centroid}}(\mathbf{X})
= \mathrm{Percentile}_{90}(D).
\end{equation}

\noindent\textbf{Complexity.}
$\mathcal{O}(n d)$\\

\subsubsection{Span (Medoid)}
\noindent\textbf{Intuition.}
This measure takes the mean distance from all samples to the medoid, 
i.e., the sample whose average distance to all other samples is minimal \citep{coxDirectedDiversityLeveraging2021}. Higher values indicate a higher diversity.
\citet{coxDirectedDiversityLeveraging2021} introduce it under the name Sparseness; we label it Span (Medoid) to pair it with Span (Centroid) (§\ref{app:tax:spancentroid}). Note that the two differ in both the reference point and the aggregation: Span (Centroid) takes a percentile of distances to a mean-based reference, whereas this measure takes a mean of distances to a median-like reference.

\noindent\textbf{Computation.}
First, the medoid is identified as:
\begin{equation}
m = \arg\min_{x_i \in \mathbf{X}} 
\sum_{j=1}^{n} d_{ij}.
\end{equation}
Compute the mean distance of all samples to this medoid:
\begin{equation}
\mathrm{Span}_{\mathrm{medoid}}(\mathbf{X})
= \frac{1}{n} \sum_{i=1}^{n} d(\emb{i}, \emb{m}).
\end{equation}

\noindent\textbf{Complexity.}
$\mathcal{O}(n^2 d)$ (pairwise distance matrix required to identify medoid)\\

\subsubsection{Cluster Inertia}
\noindent\textbf{Intuition.}
In physics, the \emph{moment of inertia} describes how difficult it is to change an object's rotational state.
Objects with more mass distributed farther from the axis of rotation have a larger moment of inertia.
By analogy, embeddings lying farther from their cluster centroid contribute more to cluster inertia.
A higher inertia value indicates that samples are scattered farther from their cluster centroids, indicating a more diverse dataset.

\noindent\textbf{Computation.}
Given a dataset $\mathbf{X}$, 
partition the embeddings into $K$ clusters $\{C_1, \ldots, C_K\}$ using $K$-means \citep{yangMeasuringDataDiversity2025, duBoostingDialogResponse2019}.
Let $\boldsymbol{\mu}_j$ denote the centroid of cluster $C_j$.
The Cluster Inertia is defined as the total within-cluster sum of squared distances:
\begin{equation}
\mathrm{Inertia}(\mathbf{X})
= \sum_{j=1}^{K} \sum_{\emb{i} \in C_j} 
  \|\emb{i} - \boldsymbol{\mu}_j\|^2.
\end{equation}

Note that unlike most other measures in our taxonomy, Cluster Inertia is not deterministic because K-means is sensitive to its random initialization.

\noindent\textbf{Complexity.}
$\mathcal{O}(n K d \, t)$,  
where $t$ is the number of K-means iterations until the cluster algorithm converges (i.e., the cluster assignments stabilize).
\\

\subsubsection{Log-Determinant Diversity (LDD)}
\noindent\textbf{Intuition.}
This measure quantifies dataset diversity as the volume spanned by the sample embeddings in feature space: the determinant of a kernel matrix equals the squared volume of the parallelepiped spanned by the underlying feature vectors \citep{kulesza2012determinantal}, and this volume was proposed as a dataset diversity measure by \citet{wang2025diversity}. The more similar the embeddings are to one another, the smaller this volume. Higher values thus indicate greater diversity. %

\noindent\textbf{Computation.}
First, construct the kernel matrix $\mathbf{K}$ with elements $k_{ij} = \cos(\emb{i}, \emb{j})$. Since cosine similarity is equivalent to the inner product between L2-normalized embeddings, this kernel matrix is positive semidefinite. More generally, other positive semidefinite kernels can also be used. The measure is defined as:
\begin{equation}
\mathrm{LDD}(\mathbf{X})
= \log \det (\mathbf{K} + \epsilon \mathbf{I}),
\end{equation}
Since the cosine kernel matrix has rank at most $\min(d,n)$, $\mathbf{K}$ is singular whenever $n > d$, and its log-determinant is then $-\infty$; we therefore add a term $\epsilon \mathbf{I}$ (with $\epsilon = 10^{-6}$) to shift the eigenvalues by $\epsilon$, following \citet{yangMeasuringDataDiversity2025}.

\noindent\textbf{Complexity.}
$\mathcal{O}(n^2 d + n^3)$ (kernel construction, then Cholesky factorization)
\\

\subsection{Distribution-based Measures}
This family reads global properties of how the embedded dataset is distributed---over the eigenvalue spectrum of a similarity matrix, over the bins of a discretized projection, or over the elements themselves. Most measures in this family are interpretable as an effective number of distinct elements or modes, several of them obtained through an entropy.

\subsubsection{Vendi Score} %
\noindent\textbf{Intuition.}
Inspired by ecology, \citet{friedman2022vendi} propose this measure, which quantifies diversity as the effective number of unique elements in a dataset, 
derived from the entropy of the eigenvalue spectrum of a similarity matrix \citep{mironov2024measuring, limbeck2024metric}. 
It achieves its maximum value ($n$) when all samples are orthogonal (fully dissimilar) and its minimum value ($1$) when all samples are identical.

\noindent\textbf{Computation.}
First, construct the similarity matrix $\mathbf{S} = [s_{ij}]$.  
Let $\tilde{\mathbf{S}} = \mathbf{S} / \mathrm{tr}(\mathbf{S})$ be its normalized form. 
The eigenvalues $\lambda_1, \ldots, \lambda_n$ of $\tilde{\mathbf{S}}$ are non-negative and sum to 1, so that they can be treated as a probability distribution
and their entropy can be calculated.
The Vendi Score is then defined as the exponential of the Shannon entropy of the eigenvalues:
\begin{equation}
\mathrm{Vendi}(\mathbf{X})
= \exp\!\left(-\sum_{i=1}^{n} \lambda_i \log \lambda_i \right).
\end{equation}

\noindent\textbf{Complexity.}
Generally $\mathcal{O}(n^2 d + n^3)$ for the eigenvalue decomposition of the $n \times n$ similarity matrix.
However, when the embedding dimension $d$ is much smaller than the sample size ($d \ll n$), it reduces the complexity to $\mathcal{O}(n d^2)$.
\\

\subsubsection{Rényi Kernel Entropy (RKE)}
\noindent\textbf{Intuition.}
This measure quantifies dataset diversity through the information-theoretic entropy of the similarity matrix in kernel space.
Proposed by \citet{jalali2023information} and further theoretically analysed by \citet{mironov2024measuring}, it estimates the effective number of modes or clusters in the data distribution by computing entropy of the eigenvalues of the normalized similarity matrix.
More uniformly distributed samples yield higher entropy (greater diversity).

\noindent\textbf{Computation.}
First, construct a normalized kernel matrix $\mathbf{S}$ with entries $S_{ij} = \frac{1}{n} s_{ij}$, where $[s_{ij}]$ is a positive semi-definite kernel with unit diagonal ($s_{ii} = 1$). Since the kernel is positive semi-definite with unit diagonal, $\mathrm{tr}(\mathbf{S}) = 1$, so its eigenvalues $\lambda_1, \ldots, \lambda_n$ are non-negative and sum to one, and can be treated as a probability distribution over spectral modes. The Rényi Kernel Entropy of order $\alpha > 0$ is defined as:
\begin{equation}
\mathrm{RKE}_{\alpha}(\mathbf{X})
= \frac{1}{1-\alpha} \log \!\left(\sum_{i=1}^{n} \lambda_i^{\alpha}\right).
\end{equation}
The most common case is $\alpha = 2$, for which the score admits a closed form that avoids eigendecomposition:
\begin{equation}
\mathrm{RKE}_2(\mathbf{X})
= - \log \!\big( \| \mathbf{S} \|_F^2 \big)
= - \log \!\Big( \frac{1}{n^2} \sum_{i,j} s_{ij}^2 \Big).
\end{equation}
where $\|\cdot\|_F$ is the Frobenius norm. Intuitively, low pairwise similarity (small $\|\mathbf{S}\|_F^2$) corresponds to a high RKE score, indicating a dataset with many distinct regions or modes.

\noindent\textbf{Complexity.}
$\mathcal{O}(n^2 d)$ for kernel construction and $\mathcal{O}(n^2)$ for Frobenius norm computation
(for $\alpha = 2$).  
More generally, computing the full eigenvalue spectrum has complexity $\mathcal{O}(n^3)$.
\\

\subsubsection{DCScore}
\noindent\textbf{Intuition.}
This measure computes dataset diversity from a classification perspective, interpreting diversity as the degree to which each sample can be distinguished from others \citep{zhuMeasuringDiversitySynthetic2025}.  
It views diversity evaluation as an $n$-class classification task, where each sample is treated as its own class. 
If samples are highly distinct, each is easily classified into its own category; if samples are similar, they are likely to be confused.  
A higher value indicates greater diversity.

\noindent\textbf{Computation.}
First, construct a kernel matrix $\mathbf{K} \in \mathbb{R}^{n \times n}$ with entries
\[
K[i,j] = \mathrm{Kernel}(\emb{i}, \emb{j}),
\]
using an inner-product or RBF kernel.  
Then apply a softmax-based classification function to obtain the class probability matrix $\mathbf{P}$:
\begin{equation}
P[i,j] = \frac{\exp\!\left(K[i,j] / \tau\right)}
{\sum_{k=1}^{n} \exp\!\left(K[i,k] / \tau\right)},
\end{equation}
where $\tau$ is a temperature hyperparameter controlling classification resolution.  
Finally, the diversity score is computed as the trace of $\mathbf{P}$, i.e., the sum of probabilities that each sample is correctly classified into its own class:
\begin{equation}
\mathrm{DCScore}(\mathbf{X}) = \mathrm{tr}(\mathbf{P}) = \sum_{i=1}^{n} P[i,i].
\end{equation}
The score ranges from $1$ (all samples identical) to $n$ (all samples distinct). 
Smaller $\tau$ increases sensitivity to smaller differences, while larger $\tau$ smooths classification probabilities.

\noindent\textbf{Complexity.}
$\mathcal{O}(n^2 d + n^2)$ for general kernels (pairwise similarity + softmax summarization).
\\

\subsubsection{Bins Entropy}
\noindent\textbf{Intuition.}
This measure quantifies how evenly sample embeddings are distributed across a discretized low-dimensional projection of the embedding space \citep{coxDirectedDiversityLeveraging2021}, by projecting the high-dimensional embeddings (e.g., with UMAP) into a 2D grid. 
Datasets whose samples occupy many bins with similar frequencies exhibit higher entropy, indicating greater diversity. 
\citet{yangMeasuringDataDiversity2025} apply the same idea of taking the Shannon entropy over bin counts, but discretize via K-means clustering rather than a grid over a projection. We follow \citet{coxDirectedDiversityLeveraging2021} and use UMAP throughout.

\noindent\textbf{Computation.}
Given embeddings $\{\emb{1}, \ldots, \emb{n}\}$ projected into a 2D space $\mathbf{Y} = \mathrm{UMAP}(\mathbf{E})$,  
divide the projected space into a $B_x \times B_y$ grid of bins $\{b_1, \ldots, b_{B}\}$, where $B = B_x B_y$.  
Let $f_b$ denote the normalized frequency (proportion) of points in bin $b$. %
The Shannon entropy is then computed as:
\begin{equation}
\begin{aligned}
\mathrm{Entropy}_{\mathrm{bins}}(\mathbf{X})
= - \sum_{b=1}^{B} f_b \log f_b,\\
\quad \text{where } f_b = \frac{n_b}{n},\; \sum_b f_b = 1.
\end{aligned}
\end{equation}
Optionally, normalize by $\log(B)$ to bound the measure between $[0,1]$:
\begin{equation}
\mathrm{Entropy}_{\mathrm{norm}} = \frac{\mathrm{Entropy}_{\mathrm{bins}}}{\log(B)}.
\end{equation}
Note that UMAP is a stochastic algorithm, so different runs can produce different results.

\noindent\textbf{Complexity.}
$\mathcal{O}(n^{1.14} d + B)$  
(Projection + histogram binning; $d$ = embedding dimension, $B$ = number of bins.)
\\

\subsubsection{MagArea (Metric Space Magnitude Area)}
\noindent\textbf{Intuition.}
This measure reads the \emph{magnitude} of the metric space---a scalar that increases as samples become more mutually dissimilar---and reports it as an effective number of distinct samples \citep{limbeck2024metric}. 
Unlike other measures in this family, magnitude is not computed at a single resolution: the similarity matrix is parameterized by a distance-scaling parameter $\tau$, and MagArea aggregates the magnitude across a range of $\tau$, so that diversity is captured jointly at fine and coarse resolutions. A higher value indicates greater diversity.

\noindent\textbf{Computation.}
First, define the similarity kernel $\mathbf{Z}_{\tau}$ parameterized by scale $\tau > 0$:
\begin{equation}
Z_{ij}(\tau) = \exp(-\tau \, d(\emb{i}, \emb{j})),
\end{equation}
The \emph{magnitude} of the metric space at scale $\tau$ is:
\begin{equation}
\mathrm{Mag}(\tau)
= \mathbf{1}^\top \mathbf{Z}_{\tau}^{-1} \mathbf{1},
\end{equation}
where $\mathbf{1}$ is the all-ones vector.  
The \textbf{MagArea} is defined as the integral of this magnitude function over a range of scales $T = [0, t_{\mathrm{cut}}]$ \citep[Def.~3.3]{limbeck2024metric}:
\begin{equation}
\mathrm{MagArea}(\mathbf{X})
= \int_{0}^{t_{\mathrm{cut}}} \mathrm{Mag}(\tau)\, d\tau.
\end{equation}
The upper limit is necessary: as $\tau \to \infty$, $\mathbf{Z}_{\tau} \to \mathbf{I}$ and $\mathrm{Mag}(\tau) \to n$, so the integral over $[0, \infty)$ diverges. Following \citet{limbeck2024metric}, we set $t_{\mathrm{cut}}$ to the median of the per-dataset convergence scales, where the convergence scale is the value of $\tau$ at which the magnitude function reaches $95\%$ of $n$; it is determined jointly across the five datasets of a sweep so that their magnitude functions are integrated over a common interval. 
In practice, the integral is approximated by evaluating $\mathrm{Mag}(\tau)$ at $n_{ts} = 30$ evenly spaced scales in $[0, t_{\mathrm{cut}}]$ and summing with trapezoidal integration; we report the resulting area divided by $t_{\mathrm{cut}}$, i.e.\ the mean of the magnitude function over the interval, which is the default behaviour of the reference implementation and does not affect the ordering within a sweep.

\noindent\textbf{Complexity.}
$\mathcal{O}(n_{ts}\, n^3)$, where $n_{ts}$ is the number of sampled $\tau$ values
: at each scale, evaluating $\mathrm{Mag}(\tau)$ requires a Cholesky factorization of the $n \times n$ similarity matrix $\mathbf{Z}_\tau$ followed by a triangular solve, rather than an explicit matrix inversion \citep{limbeck2024metric}. The pairwise distance matrix is computed once and reused across scales.
\\

\subsection{Graph-based Measures}
This family represents the dataset as a fully connected, weighted graph whose nodes are samples and whose edge weights are pairwise distances. Diversity is then measured from the structure of that graph.

\subsubsection{MST Dispersion}
\noindent\textbf{Intuition.}
Originally adapted from phylogenetics and ecology, MST Dispersion measures diversity by constructing the Minimum Spanning Tree (MST) of the embedding graph \citep{coxDirectedDiversityLeveraging2021, atwalPrivacyPreservingSyntheticReview2025}.  

\noindent\textbf{Computation.}
First, construct a fully connected, weighted graph %
where each node corresponds to a sample and each edge weight is the pairwise distance:
\[
d_{ij} = d(\emb{i}, \emb{j}).
\]
Compute the Minimum Spanning Tree $\mathrm{MST}(G)$ that connects all $n$ nodes 
with the smallest possible total edge weight.
The MST Dispersion is defined as the total weight of the MST's edges:
\begin{equation}
\mathrm{MSTDispersion}(\mathbf{X})
= \sum_{(i,j) \in \mathrm{MST}(G)} d_{ij}.
\end{equation}

\noindent\textbf{Complexity.}
$\mathcal{O}(n^2 d)$ for pairwise distance computation and 
$\mathcal{O}(n^2 \log n)$ for MST construction (e.g., Kruskal’s algorithm).

\subsubsection{Graph Entropy}
\noindent\textbf{Intuition.}
This measure takes dataset diversity as the uncertainty of pairwise distances between samples represented as a fully connected graph \citep{yuCanDataDiversity2022}.  
Each node corresponds to an embedding, and each edge weight reflects distance between two nodes.  
By computing the entropy of normalized edge weights, %
it captures how evenly pairwise distances are distributed across the dataset.
Higher entropy indicates a more uniformly distributed set of embeddings, suggesting higher dataset diversity.

\noindent\textbf{Computation.}
First form a complete weighted graph where vertices correspond to samples and edge weights are cosine distances $d_{ij} = 1 - \cos(\emb{i}, \emb{j})$.
For each node $x_i$, normalize its pairwise distances to obtain a local distance distribution:
\begin{equation}
f_i(d_{ij}) = \frac{d_{ij}}{\sum_{k=1}^{n} d_{ik}}.
\end{equation}
The local distance entropy of node $x_i$ is then:
\begin{equation}
I(x_i) = -\sum_{j=1}^{n} f_i(d_{ij}) \log f_i(d_{ij}),\footnote{The $j=i$ term vanishes since $d_{ii}=0$ implies $f_i(d_{ii})=0$; we adopt the standard convention $0\log 0 := 0$.}
\end{equation}
and the overall graph entropy of the dataset is obtained by aggregating across all nodes:
\begin{equation}
\mathrm{GraphEntropy}(\mathbf{X})
= \sum_{i=1}^{n} I(x_i).
\end{equation}

Note that since the measure sums $I(x_i)$ over all $n$ nodes, Graph Entropy grows with n and is not comparable across datasets of different sizes.

We categorize it as a graph-based measure, following \citet{yuCanDataDiversity2022}'s description and naming. However, note that a complete graph on all embedding pairs is equivalent to a distance matrix: for each sample, the Shannon entropy is computed over its normalized distances to all others. These entropies are then summed across samples. This measure could therefore arguably also be categorized as a distribution-based measure.

\noindent\textbf{Complexity.}
$\mathcal{O}(n^2 d)$ (pairwise distance computation and per-node normalization)
\\

\subsection{Surveyed but Not Benchmarked}\label{app:tax:excluded}

We surveyed ten further embedding-based measures that we do not include in \textsc{EmbDivBench}, either because they reduce to measures we already evaluate, because they are computationally intractable at our dataset size, or because they require an external reference that our controlled datasets do not provide: \textbf{KNN Distance} \citep{yangMeasuringDataDiversity2025}, \textbf{Vendi-$q$} \citep{pasarkar2023cousins}, \textbf{Species}$(q)$ \citep{leinsterMeasuringDiversityImportance2012}, \textbf{Functional Diversity} \citep{karsdorpIntroducingFunctionalDiversity}, \textbf{Density} \citep{yangMeasuringDataDiversity2025}, the dataset-level \textbf{Determinantal Point Process} score \citep{wilhelmPracticalDiversifiedRecommendations2018}, \textbf{Hamiltonian Diversity} \citep{huHamiltonianDiversityEffectively2024}, \textbf{\#Circles} \citep{xieHowMuchSpace2023}, \textbf{NovelSum} \citep{yangMeasuringDataDiversity2025} and \textbf{Sem-Ent} \citep{hanMeasuringImprovingSemantic2022}.

\section{Benchmark Construction Details}
\label{app:benchmark}

\begin{table}[!htbp]
  \centering
  \label{tab:balance-conditions}
  \small
  \setlength{\tabcolsep}{0pt}
  \renewcommand{\arraystretch}{1.15}
  \resizebox{\linewidth}{!}{
  \begin{tabular}{@{}l @{\hspace{1.4em}} l@{}}
  \toprule
  \textbf{Distribution} & \textbf{Description} \\
  \midrule
  Uniform          & Equal counts per category \\
  Slightly skewed  & Top 20\% categories take 40\% of points \\
  Mildly skewed    & Top 20\% categories take 60\% of points \\
  Zipfian          & Zipfian decay over categories \\
  Strongly skewed  & Top 1 category takes 50\%; next 4 take 30\% \\
  \bottomrule
  \end{tabular}
  }
    \caption{The five balance levels, ordered from most balanced (top) to least balanced (bottom).}
  \label{tab:balance}
\end{table}

\begin{figure*}[!p]
    \centering
    \includegraphics[width=0.9\textwidth,height=0.72\textheight,keepaspectratio]{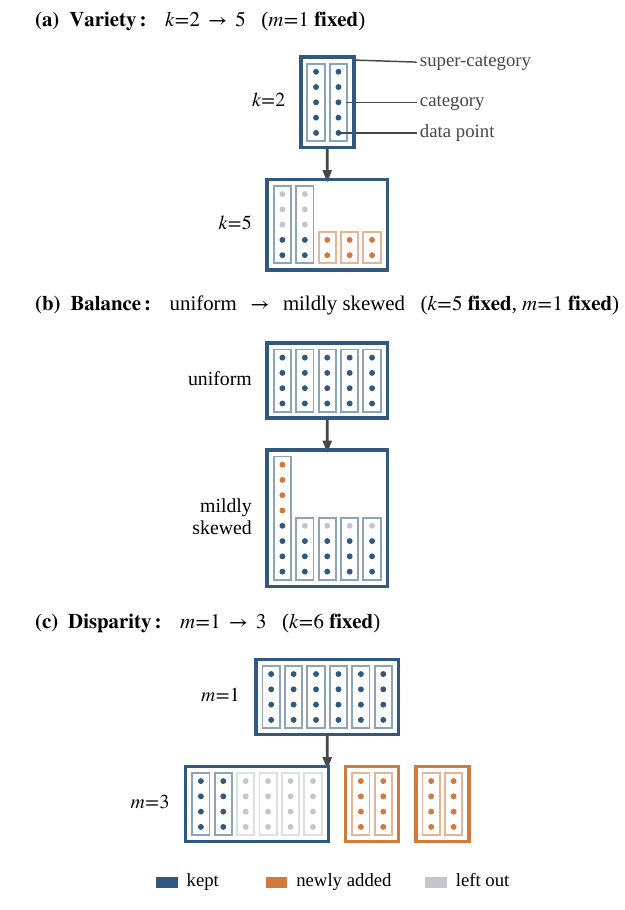}
    \caption{Incremental construction across the three components, illustrated with small example values. Each dataset is a hierarchy of \emph{super-categories} (thick outer box) containing \emph{categories} (thin inner column) containing \emph{data points} (dot). The total dataset size $n$ is conserved across levels in every panel. \textbf{Variety} ($k{=}2{\to}5$, with $m{=}1$ fixed): each kept category's data points are truncated to a prefix, and 3 new categories are added. \textbf{Balance} (uniform $\to$ mildly skewed; $k{=}5$ and $m{=}1$ both fixed): the category set is held constant; data points are redistributed from low-frequency to high-frequency categories. \textbf{Disparity} ($m{=}1{\to}3$, with $k{=}6$ fixed): the kept super-category retains a fixed-order prefix of 2 of its 6 categories (the other 4 are left out), and 2 new super-categories are added. Blue marks elements kept from the lower diversity level, orange marks elements newly added, and grey marks elements that were in the lower level but are left out at the higher level; this colour coding applies at all three levels of the hierarchy.}
    \label{fig:incremental}
\end{figure*}

\paragraph{Incremental construction.} Along every component, moving from one level to the next \emph{extends} the previous configuration rather than replacing it, so observed measure responses cannot be attributed to swapping categories between levels. Fig.~\ref{fig:incremental} illustrates the three components with one example transition per component, using small values for visual clarity. Take \textbf{variety} as an example: the $k{=}2$ category set is a strict subset of the $k{=}5$ set, and the data points retained from each shared category are exactly the first 2 of those used at $k{=}2$ (since $n$ is fixed and the per-category budget shrinks from 5 to 2). On the \textbf{balance} component, the category set is held constant across all conditions and only the allocation of data points changes. On the \textbf{disparity} component, the $m{=}1$ super-category set is a strict subset of the $m{=}3$ set, and within each retained super-category, the categories are selected by a fixed-order prefix---so changes in the $k{=}6$ category set between adjacent $m$ levels are driven by super-category extension and the fixed $k$ budget, not by random replacement.

\subsection{Simulated Tier Configuration}
\label{app:gmm}

We sweep five embedding dimensionalities $d \in \{256, 384, 1024, 2048, 4096\}$ to characterize dimension-dependent measure behaviour; results in the main body use $d{=}384$.

\paragraph{Hierarchical Gaussian mixture.} Super-category centres are drawn from $\mathcal{N}(\mathbf{0}, \sigma_{\textrm{super}}^2 I_d)$ with $\sigma_{\textrm{super}} = 5\sqrt{d}$; category centres from $\mathcal{N}(\mathbf{s}_{m(i)}, \sigma_{\textrm{category}}^2 I_d)$ with $\sigma_{\textrm{category}} = \sqrt{d}$; and each element of category $i$ from $\mathcal{N}(\mathbf{c}_i, \sigma_{\textrm{element}}^2 I_d)$ with $\sigma_{\textrm{element}} = 0.5$ (constant in $d$). The two larger scales grow with $\sqrt{d}$ because pairwise distances between isotropic Gaussian draws scale with $\sqrt{d}$; this keeps the relative separation of super-category centres, category centres, and intra-category clouds comparable across dimensionalities. The assignment of category $i$ to a super-category is deterministic, $m(i) = i \bmod m$, so categories are distributed as evenly as possible across the $m$ super-categories.

\subsection{Natural Text Tier Configuration}
\label{app:wiki-pool}

Construction has two stages: choosing the domain/topic structure by farthest-point sampling (FPS) over label embeddings (\S\ref{app:wiki-fps}), and collecting sentences for each chosen topic by adaptively crawling its Wikipedia category (\S\ref{app:wiki-pool-crawl}). Tab.~\ref{tab:hierarchy} shows examples of the resulting domain/topic structure.

\paragraph{Topic selection.} We build the candidate pool from a one-time crawl of Wikipedia's category hierarchy, accessed through the live MediaWiki API (via the \texttt{Wikipedia-API} Python package\footnote{\url{https://pypi.org/project/Wikipedia-API/}}) on 24 April~2026. The top level (``Life'', ``Concepts'', ``People'', etc.) is excluded as too broad to be a meaningful categorical unit. Domains (e.g.\ \emph{Humanities}, \emph{Engineering disciplines}) correspond to super-categories in the benchmark's notation, and topics within those domains (e.g.\ \emph{Anthropology}, \emph{Materials science}) correspond to categories. A topic is retained in the pool only if its Wikipedia category has at least 30 direct article pages, which targets topics with enough material to support the largest per-topic sentence count we need. After deduplicating topic labels (a small fraction of topics appear under multiple domains; we keep the one with the most direct pages), the candidate pool contains 191 domains, each with at least 5 suitable topics. Because the live API returns article state at request time, re-running the construction pipeline at a later date will yield a similar but not identical pool as Wikipedia content evolves; the released benchmark archive contains the exact sentences used in our experiments, so reproducing the results in this paper does not require re-crawling.

\begin{table*}[!htbp]
\centering
\small
\begin{tabular}{l l p{8.5cm}}
\toprule
\textbf{Domain} & \textbf{Topic} & \textbf{Example sentence} \\
\midrule
\multirow{3}{*}{Government by country}
  & \multirow{2}{*}{National legislatures}
  & A general election is an electoral process to choose most or all members of a governing body at the same time. \\
  & & After being elected the Prime Minister appoints the cabinet ministers and announces them to the Riksdag. \\
\cmidrule{2-3}
  & Parliaments by country
  & Article 16.4.2\textdegree{} requires that the newly elected Dáil Éireann must convene no later than 30 days after the election. \\
\midrule
Humanities & Anthropology
  & A horse culture is a tribal group or community whose day-to-day life revolves around the herding and breeding of horses. \\
\midrule
Engineering disciplines & Materials science
 & Air and vacuum have high reluctance, while easily magnetized materials such as soft iron have low reluctance. \\
\bottomrule
\end{tabular}
\caption{Examples from the Wikipedia Domains/Topics hierarchy. Sentences within the same topic share subject matter; sentences within the same domain but different topics are related but distinguishable; sentences from different domains are semantically distant.}
\label{tab:hierarchy}
\end{table*}

\subsubsection{Farthest-Point Sampling (FPS) of Domains and Topics}
\label{app:wiki-fps}

Some domains in the candidate pool can be semantically related (e.g., several concern history or politics). Selecting domains uniformly at random would risk drawing a set concentrated in one region of the label-embedding space, which would undermine the structural assumption behind the natural text tier: that different domains are semantically distant while topics within a domain are related but distinguishable. We therefore use farthest-point sampling (FPS), a greedy procedure that starts from one domain and repeatedly adds the candidate farthest from the already-selected set, thereby spreading the selection across the label-embedding space.

\paragraph{Procedure.} Domain and topic labels are encoded once with Qwen3-Embedding-8B and L2-normalized; because the embeddings are unit-norm, Euclidean FPS is monotonically equivalent to cosine FPS. 
We apply FPS first over domain labels and then, within each chosen domain, apply the same procedure over its topic labels. Given its starting label, FPS is deterministic; the only source of randomness per run is therefore the choice of the starting label, which is drawn under the run seed. The full domain/topic selection is thus reproducible from the run seed alone.

\paragraph{Two selection settings.} Variety and balance both hold the domain structure fixed at $m=10$ and use at most $k=50$ topics, so they share a single selection of 10 domains with 5 topics each. Disparity instead varies $m$ from 10 to 50 at fixed $k=50$ and thus requires up to 50 domains; we run a separate domain FPS for it with different seeds and, at each level, select the required number of topics per domain (one or more). Both selections use the same FPS procedure described above.

\subsubsection{Sentence Pool Construction}
\label{app:wiki-pool-crawl}

\paragraph{Adaptive expansion schedule.} For each chosen topic, we build a sentence pool by crawling its Wikipedia category in up to 5 rounds with linearly increasing parameters: descent depth $1 \to 3$, articles inspected per round $4{,}000 \to 20{,}000$, and per-article sentence cap $5 \to 40$. A round only fires when the pool has not yet reached the required size; in the typical case the first round (depth 1, cap 5) suffices, and later rounds are reserved for sparser topics. This keeps the dominant sampling regime close to the ``shallow, low cap'' setting (i.e., the shallow descent avoids semantic drift from deep subcategory descent and the low per-article cap prevents any single article dominating a topic's pool), while still allowing thin topics to reach their target pool size.

\paragraph{Crawl mechanics.} Within a round, articles are collected breadth-first from the root category: members of the current category are inspected in shuffled order, and we descend into a subcategory only when the current depth has not been exhausted and the article budget has not yet been hit. For each article we extract its full body text, split it into sentences with a simple end-punctuation regex, shuffle the sentences, and add them to the pool subject to the length filter (100--300 characters) and the per-article cap. Sampling sentences in shuffled order rather than as a leading prefix avoids overrepresenting the stylized opening sentences characteristic of Wikipedia articles.

\subsection{Hyperparameter Settings}

We list the hyperparameter settings for all 21 measures in Tab.~\ref{tab:hyperparams}. 

\begin{table}[!htbp]
\centering
\small
\begin{tabular}{@{}lp{3.9cm}@{}}
\toprule
\textbf{Measure} & \textbf{Hyperparameters} \\
\midrule
\multicolumn{2}{@{}l@{}}{\textit{Distance-based}} \\
Mean Pairwise Dist.  & --- \\
Sum Pairwise Dist.   & --- \\
Energy               & $\gamma=1.0$, $\epsilon=10^{-12}$ \\
Diameter             & --- \\
Bottleneck           & --- \\
Sum Diameter         & no division by $n$ \\
Sum Bottleneck       & no division by $n$ \\
Chamfer Dist.        & --- \\
\addlinespace
\multicolumn{2}{@{}l@{}}{\textit{Geometry-based}} \\
Convex Hull Vol. (2D) & 2D UMAP projection, \texttt{random\_state} $=42$ \\
Span (Centroid)      & 90th percentile \\
Span (Medoid)        & --- \\
Cluster Inertia      & $K=200$ \\
Geo. Mean of Std.    & no distance metric \\
Log-Determinant      & cosine kernel, $\tau=1.0$, normalized, $\epsilon=10^{-6}$, Cholesky \\
\addlinespace
\multicolumn{2}{@{}l@{}}{\textit{Distribution-based}} \\
Vendi Score          & $q=1.0$, normalized, dual form \\
Rényi Kernel Ent.    & $\alpha=2.0$, cosine kernel, $\tau=1.0$, normalized, $\epsilon=10^{-12}$ \\
DCScore              & cosine kernel, $\tau=1.0$, normalized \\
Bins Entropy (UMAP)  & $5\times5$ grid, UMAP projection, effective normalization \\
MagArea              & $n_{\text{ts}}=30$ scales \\
\addlinespace
\multicolumn{2}{@{}l@{}}{\textit{Graph-based}} \\
MST Dispersion       & --- \\
Graph Entropy        & --- \\
\bottomrule
\end{tabular}
\caption{Hyperparameter settings for all 21 measures. ``---'' means the measure takes no further parameters. Measures differ in whether they operate on cosine distances or on a cosine-similarity kernel. More details can be found in our repository at \url{https://github.com/nlpsoc/EmbDivBench}}
\label{tab:hyperparams}
\end{table}

\section{Topic Label Noise Audit for the Natural Text Tier}
\label{app:noise}

The natural text tier uses Wikipedia topic labels as a proxy for sentence-level semantics (\S\ref{sec:wiki}), so not every sentence taken from an article is necessarily a good representation of the article's topic. We therefore manually audit the Wikipedia topic labels.

\paragraph{Setup.} We draw a stratified sample of 10 sentences from every dataset of the three components: for variety, the datasets with $k \in \{10, 20, 30, 40, 50\}$ topics ($m=10$ fixed); for balance, the five distribution conditions ($k=50$, $m=10$ both fixed); and for disparity, the datasets with $m \in \{10, 20, 30, 40, 50\}$ domains ($k=50$ fixed), each across all 5 seeds, yielding $3 \times 5 \times 5 \times 10 = 750$ sentences. 
One of the authors checks each sentence against the full topic pool of its source dataset and annotates it with a binary label: \textsc{match} if the sentence supports its assigned Wikipedia topic, \textsc{mismatch} otherwise. The noise rate is the proportion of \textsc{mismatch} over all 750 annotated sentences.

\paragraph{Judgement criteria.} A sentence is a \textsc{mismatch} when it provides no usable evidence for its assigned topic: it is about a different topic in the pool, it is too generic to point anywhere (e.g., a transition or background sentence), or the sentence splitter returned an incomplete snippet. Besides exact matches, two other cases are also counted as \textsc{match}. First, a sentence that is about its assigned topic but would also fit a neighbouring topic in the pool is still a match, because it sits in the right region of the embedding space, and the levels of each component are separated by wide intervals ($k$ and $m$ both move in steps of 10), so an ambiguous placement does not damage the structure of our designed benchmark. Second, a sentence that is about its topic only if one knows some external fact is also a \textsc{match}, because embedding models are pre-trained on broad corpora and have access to exactly the world knowledge human annotators may lack. This rule is, however, easier to state than to apply (see below \emph{Annotation difficulty}). Tab.~\ref{tab:noise-examples} illustrates both labels with examples.

\begin{table*}[p]
\centering
\small
\renewcommand{\arraystretch}{1.3}
\begin{tabular}{@{}p{0.46\linewidth} p{0.17\linewidth} p{0.31\linewidth}@{}}
\toprule
\textbf{Sentence} & \textbf{Assigned topic} & \textbf{Comment} \\
\midrule
\multicolumn{3}{@{}l}{\textbf{\textsc{match}} \emph{--- exact matches}} \\
\addlinespace[3pt]
He also contributed a foreword to white supremacist commentator Stefan Molyneux's book \emph{Against the Gods}.
& New Atheism
& The topic is explicit in the sentence itself. \\
\addlinespace[6pt]

\multicolumn{3}{@{}l}{\textbf{\textsc{match}} \emph{--- neighbouring topics}} \\
\addlinespace[3pt]
In German, for example, the pentagraph \emph{tzsch} represents the [t\textesh] sound of the English digraph \emph{ch}, and indeed is found in the English word \emph{Nietzschean}.
& Orthography
& \emph{Orthography} and \emph{Linguistic terminology} both make sense, but they lie close enough in the embedding space. \\
\addlinespace[6pt]

\multicolumn{3}{@{}l}{\textbf{\textsc{match}} \emph{--- requires background knowledge}} \\
\addlinespace[3pt]
Quickly was founded in California in 1996 and started its trademark licensing program at the same time.
& Tea culture
& Resolvable only by recognizing Quickly as a Taiwanese bubble tea chain. \\
\addlinespace[3pt]
George married Edith Jane Stewart Bayley (died in England of a painful internal cancer 8 June 1876) on 18 December 1854.
& Government of Australia
& Requires knowing the named subject is an Australian politician; reads as a generic biographical note otherwise. \\
\addlinespace[6pt]

\multicolumn{3}{@{}l}{\textbf{\textsc{mismatch}} \emph{--- different topic}} \\
\addlinespace[3pt]
In 1938 he broke the Turkish national record, with a time of 49.0 that was not surpassed for seven years, while winning the 1938 Balkan Games 400\,m title in Belgrade.
& Jewish emigration from Nazi Germany
& The article covers the subject's emigration, but this sentence reads as another topic in the pool \emph{Sports records and statistics}. \\
\addlinespace[3pt]
Meats cooked at high temperatures, especially above 300\,\textdegree F (149\,\textdegree C) (as in grilling or pan frying), or that are cooked for a long time tend to form more HCAs.
& Animal products
& Describes preparation rather than the product itself; closer to another topic \emph{Cooking techniques}. \\
\addlinespace[6pt]

\multicolumn{3}{@{}l}{\textbf{\textsc{mismatch}} \emph{--- no topical cue}} \\
\addlinespace[3pt]
Gerhard Fisher studied electronics at the University of Dresden before emigrating to the United States.
& Prisons
& Biographical background; carries no cue to any topic in the pool. \\
\addlinespace[3pt]
He holds a bachelor's degree from California State University, Long Beach, and a Master of Fine Arts (MFA) from California State University, Fullerton.
& Medical journal editors
& The referent of \emph{he} is only recoverable from the surrounding context. \\
\addlinespace[6pt]

\multicolumn{3}{@{}l}{\textbf{\textsc{mismatch}} \emph{--- incomplete snippet}} \\
\addlinespace[3pt]
Cabinet Ministers Didier Budimbu - 2022-present Aime Ngoy Mukena - 2015--2022 Crispin Atama Tabe - 2012--2015 [\dots] == References ==
& Government of Iran
& Not a sentence: a list of names and a section heading picked up by the sentence splitter. \\
\bottomrule
\end{tabular}
\caption{Illustrative examples from the manual audit, grouped by the reason for each label. Examples are selected for illustration and are not proportional to the frequency of each case.}
\label{tab:noise-examples}
\end{table*}

\paragraph{Overall noise level.} Of the 750 sentences, 616 match their assigned topic and 134 are mismatched, giving a noise rate of 17.9\% (134/750; Wilson 95\% CI [15.3\%, 20.8\%]). The three components give comparable rates (variety 20.4\%, balance 15.2\%, disparity 18.0\%), and within each component the rate does not vary substantially across levels (Spearman $\rho=0.46$, $p=0.43$ for variety; $\rho=0.20$, $p=0.75$ for balance; $\rho=-0.11$, $p=0.87$ for disparity). Label noise therefore affects the five level datasets of a component to a comparable degree rather than increasingly with the diversity level, so it degrades all five datasets alike and does not act as a confound for the benchmark design. Tab.~\ref{tab:noise-per-level} breaks the counts down by component and level.

\paragraph{Annotation difficulty.} Assigning a sentence against up to 50 candidate topics is a non-trivial task, and annotator-dependent: even presented with the same sentence and the same topic pool, different annotators may perceive different topics as the closest fit, particularly where several candidates are semantically adjacent. The judgement also depends on the annotator's background knowledge rather than on the sentence alone. For instance, \emph{``Quickly was founded in California in 1996 and started its trademark licensing program at the same time.''} points clearly to tea culture for a reader who recognizes Quickly as the Taiwanese bubble tea chain brand, and nowhere at all otherwise; \emph{``GMR-2 is not an evolution of GMR-1 but rather a concurrent standard that has been developed by another group of companies.''} is clearly about satellite communication to an engineer. Where the required knowledge is missing, an annotator is more likely to judge a sentence \textsc{mismatch} than to recognize its link to the assigned topic (see the knowledge-dependent rows in Tab.~\ref{tab:noise-examples}).

\begin{table}[htbp]
\centering
\footnotesize
\setlength{\tabcolsep}{4pt}
\begin{tabular}{@{}lcccc@{}}
\toprule
Level & \textsc{match} & \textsc{mismatch} & Noise \\
\midrule
\multicolumn{4}{@{}l}{\textit{Variety}} \\
\hspace{0.6em}$k=10$ & 44 & 6 & 0.12 \\
\hspace{0.6em}$k=20$ & 40 & 10 & 0.20 \\
\hspace{0.6em}$k=30$ & 36 & 14 & 0.28 \\
\hspace{0.6em}$k=40$ & 39 & 11 & 0.22 \\
\hspace{0.6em}$k=50$ & 40 & 10 & 0.20 \\
\addlinespace
\multicolumn{4}{@{}l}{\textit{Balance}} \\
\hspace{0.6em}Uniform         & 41 & 9 & 0.18 \\
\hspace{0.6em}Slightly skewed & 46 & 4 & 0.08 \\
\hspace{0.6em}Mildly skewed   & 43 & 7 & 0.14 \\
\hspace{0.6em}Zipfian         & 40 & 10 & 0.20 \\
\hspace{0.6em}Strongly skewed & 42 & 8 & 0.16 \\
\addlinespace
\multicolumn{4}{@{}l}{\textit{Disparity}} \\
\hspace{0.6em}$m=10$ & 39 & 11 & 0.22 \\
\hspace{0.6em}$m=20$ & 42 & 8 & 0.16 \\
\hspace{0.6em}$m=30$ & 42 & 8 & 0.16 \\
\hspace{0.6em}$m=40$ & 41 & 9 & 0.18 \\
\hspace{0.6em}$m=50$ & 41 & 9 & 0.18 \\
\bottomrule
\end{tabular}
\caption{Per-level annotation counts (50 sentences per level) and noise rate (\textsc{mismatch} over annotated sentences) for the three components of the natural text tier.}
\label{tab:noise-per-level}
\end{table}

\section{Results for Robustness Check} 
\label{app:dim-robustness}
\input{figures/fig_cross_dimensions}
\input{figures/fig_cross_embedders}
\input{figures/fig_wiki_miniLM-style}

We elaborate on results for the robustness check. We display results on the natural text tier for \emph{Qwen3-Embedding-8B}\footnote{\url{https://huggingface.co/Qwen/Qwen3-Embedding-8B}} and \emph{e5-mistral-7b-instruct\footnote{\url{https://huggingface.co/intfloat/e5-mistral-7b-instruct}}} compared to \emph{MiniLM} in Fig.~\ref{fig:cross-embedders} and results for \emph{Style Embedding}\footnote{\url{https://huggingface.co/AnnaWegmann/Style-Embedding}} compared to \emph{MiniLM} in Fig.~\ref{fig:cross-embedders-style}. 

\paragraph{Qwen3 improves correlations and yields the same family ranking.} Several findings with \emph{MiniLM} hold with \emph{Qwen3} and \emph{E5}, high-performing embedding models on MTEB\footnote{Note that MTEB includes not only semantic tasks but also application-oriented tasks such as clustering and retrieval, which may impose encoding requirements beyond semantic similarity.} \cite{enevoldsenMMTEBMassiveMultilingual2025}: no measure performs perfectly on all components, the family ranking (Graph $>$ Distribution $>$ Geometry $>$ Distance) is consistent on all three embedding models, and disparity remains the hardest component. Further, most measures reach higher correlation with Qwen3 embeddings, with a few decreasing correlation on disparity.

\section{Detailed Rationale for Recommendations}
\label{app:recommendations}
On disparity, Graph Entropy scores higher than MST Dispersion in our main experiment (+0.80 vs.\ +0.68 with MiniLM; Fig.~\ref{fig:results-natural-text-d384-mean}), but this ordering does not hold across embedding models: averaged over the three semantic embedders of §\ref{sec:robust-model} (Fig.~\ref{fig:cross-embedders}), MST Dispersion reaches +0.70 against Graph Entropy's +0.61, with Graph Entropy dropping to +0.50 under Qwen3. We therefore recommend MST Dispersion for disparity as the option that holds up across embedders. Note that disparity is the hardest component for all measures (\S\ref{sec:results}), and even the recommended MST Dispersion is relatively weak on disparity compared to our variety and balance recommendations. Our disparity recommendation should therefore be read as the best currently available option rather than a reliable one, and we see measuring disparity as a pressing open problem for future work.

Overall, Graph Entropy has the highest aggregated performance scores across the three components, so if no single component is of primary interest, it is currently the best-supported single choice. As a complement, we also suggest Vendi ($q{=}1$): of the three measures responding perfectly to all three components at every simulated dimension we test (Fig.~\ref{fig:cross-dimensions}), Vendi ranks second on natural text averaged over the three semantic embedding models, behind Graph Entropy. It is also considerably cheaper to compute than Graph Entropy (Tab.~\ref{tab:cost}).

\section{Statistics for Data}
\label{app:stats}

Each dataset contains $n{=}10{,}000$ elements. The simulated tier sweeps 5 embedding dimensionalities, with 5 seeds and 15 configurations (5 levels for each of the three components) per dimensionality, giving $5 \times 5 \times 15 = 375$ datasets. The natural text tier uses 5 seeds and the same 15 configurations, giving $5 \times 15 = 75$ datasets. In total, the benchmark comprises $450$ datasets.

\section{Computational Cost and Budget}
\label{app:cost}

Embedding work---Qwen3-Embedding-8B for natural text data creation (\S\ref{app:wiki-fps}) and all-MiniLM-L6-v2 for sentence encoding for measure evaluation---was run on a single A100 GPU (79\,GB MIG slice). All other stages, namely simulated dataset generation, the Wikipedia category crawl, and diversity-measure evaluation, ran on CPU with 4--6 cores and up to 64\,GB of RAM. A full pass of measure evaluation over all 75 natural text datasets ($n{=}10{,}000$, $d{=}384$) takes approximately 25 CPU-hours; as Tab.~\ref{tab:cost} shows, per-measure cost spans more than four orders of magnitude, with MagArea alone accounting for $\sim$87\% of the total ($\sim$21.8\,h) and the remaining 20 measures together completing in $\sim$3.4\,h.

\begin{table*}[!htbp]
  \centering
  \caption{Compute cost and asymptotic complexity of measures. Relative cost is reported as a multiple of mean pairwise distance on identical inputs ($n{=}10000$, $d{=}384$); within each family, rows are ordered by ascending cost.}
  \label{tab:cost}
  \footnotesize
  \setlength{\tabcolsep}{6pt}
  \begin{tabular}{@{}l c@{\hspace{4pt}} l r l@{}}
  \toprule
  \textbf{Family} & & \textbf{Measure} & \textbf{Cost} & \textbf{Complexity} \\
  \midrule
  \multirow{8}{*}{\textbf{Distance}}
    &              & Mean Pairwise Dist.            & $1.00\times$    & $O(n^2 d)$ \\
    &              & Sum Pairwise Dist.             & $1.00\times$    & $O(n^2 d)$ \\
    &              & Energy                         & $1.02\times$    & $O(n^2 d)$ \\
    & \dotB        & Chamfer Dist.                  & $1.05\times$    & $O(n^2 d)$ \\
    &              & Diameter                       & $0.99\times$    & $O(n^2 d)$ \\
    &              & Bottleneck                     & $0.99\times$    & $O(n^2 d)$ \\
    &              & Sum Diameter                   & $1.03\times$    & $O(n^2 d)$ \\
    &              & Sum Bottleneck                 & $1.03\times$    & $O(n^2 d)$ \\
  \midrule
  \multirow{6}{*}{\textbf{Geometry}}
    &              & Convex Hull Vol. (2D)$^{\dagger}$  & $<\!0.01\times$ & $O(n^{1.14} d + n \log n)$ \\
    &              & Span (Centroid)                & $0.01\times$    & $O(n d)$ \\
    &              & Geo. Mean of Std.              & $0.01\times$    & $O(n d)$ \\
    &              & Log-Determinant                & $0.55\times$    & $O(n^3 + n^2 d)$ \\
    &              & Span (Medoid)                  & $1.03\times$    & $O(n^2 d)$ \\
    &              & Cluster Inertia$^{\ddagger}$                & $1.17\times$    & $O(n K d \cdot \text{iters})$ \\
  \midrule
  \multirow{5}{*}{\textbf{Distribution}}
    &              & Vendi Score                    & $0.03\times$    & $O(n d^2 + d^3)$ \\
    &              & DCScore                        & $0.16\times$    & $O(n^2 d)$ \\
    &              & R\'enyi Kernel Ent.            & $0.20\times$    & $O(n^2 d)$ \\
    &              & Bins Entropy (UMAP)$^{\dagger}$ & $1.93\times$    & $O(n^{1.14} d)$ \\
    &              & MagArea                        & $108.8\times$   & $O(n_{ts} \cdot n^3)$ \\
\midrule
  \multirow{2}{*}{\textbf{Graph}}
    & \dotG & Graph Entropy                  & $1.13\times$    & $O(n^2 d)$ \\
    & \dotR\,\dotB        & MST Dispersion                 & $3.49\times$    & $O(n^2 d)$ \\

  \bottomrule
  \end{tabular}\\[3pt]
  {\scriptsize Cost relative to MPD ($\approx 9.6$\,s wall-time). $^{\dagger}$ relies on a 2D UMAP projection, not directly comparable. $^{\ddagger}$Cluster inertia uses KMeans with $K{=}200$ clusters and default \texttt{sklearn} convergence (\texttt{iters} $\leq 300$). Coloured dots mark the recommendations for different components: blue for variety, green for balance and red for disparity.}
\end{table*}

\section{Intended Use and Licenses}
\label{appendix:licenses}

We document the licences and intended use of the models and datasets re-used and created in this work. All third-party resources are used in a manner consistent with their licences.

\paragraph{Sentence encoders.} We use \texttt{all-MiniLM-L6-v2} (22M parameters) \cite{wangMiniLMDeepSelfAttention2020,reimersSentenceBERTSentenceEmbeddings2019} as the primary sentence encoder for the natural text tier; it is released under the Apache~2.0 licence.\footnote{\url{https://huggingface.co/sentence-transformers/all-MiniLM-L6-v2}} To assess robustness across encoders of different scale, model family, and training objective, we additionally use three encoders. \texttt{Qwen3-Embedding-8B} (8B parameters) \citep{zhangQwen3EmbeddingAdvancing2025}, an LLM-based encoder, is released by Alibaba Cloud under the Apache~2.0 licence.\footnote{\url{https://huggingface.co/Qwen/Qwen3-Embedding-8B}} \texttt{e5-mistral-7b-instruct} (7B parameters) \citep{wangImprovingTextEmbeddings2024} is released under the MIT licence.\footnote{\url{https://huggingface.co/intfloat/e5-mistral-7b-instruct}} \texttt{AnnaWegmann/Style-Embedding} (125M parameters) \citep{wegmann-etal-2022-author}, a RoBERTa-based encoder trained for content-independent style representation rather than semantic similarity, is released under the MIT licence.\footnote{\url{https://huggingface.co/AnnaWegmann/Style-Embedding}} All four models are used strictly for sentence embedding extraction, consistent with their intended use.

\paragraph{Crawling library.} The Wikipedia crawler (\S\ref{app:wiki-pool}) uses the \texttt{Wikipedia-API} Python package\footnote{\url{https://pypi.org/project/Wikipedia-API/}}, released under the MIT licence, for category traversal and article retrieval via the live MediaWiki API. We use it strictly for read-only access to publicly available Wikipedia content, consistent with its intended use.

\paragraph{Wikipedia.} The natural text tier (\S\ref{sec:wiki}) is constructed by sampling sentences from English Wikipedia articles organized under the categorical hierarchy. Wikipedia text is dual-licensed under the Creative Commons Attribution-ShareAlike 4.0 International (CC~BY-SA~4.0) and the GNU Free Documentation License (GFDL).\footnote{\url{https://en.wikipedia.org/wiki/Wikipedia:Copyrights}} Re-use for academic research and benchmark construction is permitted; under the share-alike provision, any redistributed Wikipedia-derived corpus must be released under the same licence.

\paragraph{Simulated Data.} The simulated datasets used in the simulated tier (\S\ref{sec:simulated}) are generated by our own sampling code from hierarchical Gaussian mixtures. They contain no text, no third-party content, and no personally identifiable information. They are released together with the generation scripts under the Creative Commons Attribution 4.0 International (CC~BY~4.0) licence.

\paragraph{\textsc{EmbDivBench} Release.} The benchmark---including dataset construction scripts, pre-computed embeddings, the code to call the functions for different measures---is publicly released to support reproducibility and further research on embedding-based diversity measurement. The curated Wikipedia data is released under CC~BY-SA~4.0, inheriting the licence of the underlying Wikipedia text; the simulated datasets and source code are released under CC~BY~4.0. The intended use of \textsc{EmbDivBench} is the evaluation of embedding-based diversity measures within NLP/ML research; we do not anticipate use beyond this scope.

\section{Use of AI Assistants}
We used Claude Code and GitHub Copilot for coding tasks---primarily command lookup and occasional function generation, with all generated code reviewed and tested before use. We also used Claude for rephrasing and grammatical correction during writing.

\end{document}